\documentclass[11pt]{article}

\usepackage[utf8]{inputenc}
\usepackage{lmodern}
\usepackage[T1]{fontenc}
\usepackage[margin=1in]{geometry}
\usepackage{microtype}
\usepackage{amsmath,amssymb,amsthm}
\usepackage[colorlinks=true,citecolor=blue,linkcolor=blue,urlcolor=blue]{hyperref}
\usepackage{booktabs}
\usepackage{tabularx}
\usepackage{tikz}
\usepackage{pgfplots}
\pgfplotsset{compat=1.17}
\usetikzlibrary{arrows.meta,positioning,shapes.geometric}
\usepackage{csquotes}
\MakeOuterQuote{"}

\usepackage{changepage}
\newlength{\extralength}
\usepackage[style=authoryear,backend=biber,natbib=true,giveninits=true,uniquename=init]{biblatex}

\theoremstyle{definition}
\newtheorem{Definition}{Definition}
\theoremstyle{plain}
\newtheorem{Proposition}{Proposition}

\title{Zero-Shot Weight-Space Programming Across a Difficulty-Graded Ladder of Two-Player Games}
\author{Eduardo C. Garrido Merch\'an \\
Universidad Pontificia Comillas (IIT, DIGNUM) \\
Madrid, Spain \\
\texttt{ecgarrido@comillas.edu}}
\date{November 2026}

\begin{document}

\maketitle

\begin{abstract}
Weight-space programming sets every weight and bias of a feed-forward
network by hand, from an explicit specification of the computation the network
must perform, with no gradient descent and no training data anywhere in the
construction. How far a large language model can carry that practice, and where
it stops, has not been measured. We measure it on a difficulty-graded ladder of
two-player games running from Nim to a chess endgame, with a structurally
unrelated replicate game at every rung, and we check all ten resulting networks
against independently written oracles, six of them exhaustively over the
complete reachable state space. The constructions are exact wherever optimal
play admits a local certificate that a single forward pass can check, such as a
Grundy value, the legality of a candidate move, or an immediate checkmate. They
collapse wherever competent play requires composing many such local decisions
into a multi-move plan: the two chess endgames fall from 84.9 and 100 per cent
success at the one-ply tactic to 0.57 and 9.13 per cent in full adversarial
play. A policy carrying the same hand-set evaluation but two plies of search
beats the constructed network in every game, which locates the gap in the
architecture itself, under any comparable choice of coefficients. Six further
construction attempts, spanning three providers and four model sizes, reproduce
the pattern and show that the ceiling constrains the derivation of a compact
circuit while leaving the architecture's representational capacity intact. Read as game theory, the
ladder grades games by how much of equilibrium play a fixed circuit certifies
locally and how much has to be tabulated position by position.
\end{abstract}

\section{Introduction}

Gradient descent on data is how the field fixes the weights of a neural
network. The representation itself is indifferent to their provenance: a
feed-forward architecture with rectified-linear activations represents a
piecewise-linear function of its input once its coefficients are fixed, whatever
procedure produced them. \citet{weightspaceprogrammingblog} exploited that indifference directly.
They showed that a large language model can read and write the parameters of a
neural network as literal numbers, diagnosing what a trained network had been
trained on, repairing a reinforcement-learning policy by editing a small
fraction of its parameters, and writing new game-playing networks from scratch
for several grid-world environments and for Go on a small board. They named the
practice weight-space programming. In a companion study we applied the same
practice to supervised prediction and classification on structured datasets
\citep{garridomerchan2026tabular}; the present paper takes it to two-player
games.

What that body of work establishes is that the capability exists. What it does
not establish is how far the capability extends. Each demonstration reports a
task on which a construction succeeded, and a reader has no way to tell whether
the chosen tasks sit comfortably inside the region where the technique works or
at its edge. The quantity that remains unmeasured is the boundary itself: which
target behaviours a model can compile into weights by reasoning alone, which it
cannot, and what property of the target decides between the two cases. Nothing
in the existing evidence separates a target that is genuinely out of reach from
one that a more careful construction would have reached.

That boundary matters for three reasons. A hand-set network is auditable in a
way a trained network is not, because every coefficient has a stated derivation
and the whole object can be checked against an independent specification rather
than estimated on a held-out sample. A technique whose region of applicability
is known can be used deliberately, where one whose region is unknown can only be
tried. And the location of the boundary is informative about the models doing
the constructing, since it separates what a model can state about a computation
from what it can realize as an artifact that runs.

\begin{figure}[htbp]
\centering
\resizebox{\textwidth}{!}{%
\begin{tikzpicture}[
  node distance=0.9cm and 1.3cm,
  block/.style={draw, rounded corners, align=center, fill=gray!8, minimum width=3.3cm, minimum height=1.6cm, font=\small, text width=3.0cm},
  arr/.style={-{Latex[length=2.2mm]}, thick},
  lbl/.style={font=\scriptsize, align=center, text width=2.6cm},
]
\node[block] (spec) {Game specification: rules, state encoding, required input and output tensor shapes, one worked example from a different game, all as text};
\node[block, right=of spec] (llm) {Large language model: no gradient descent, no training data, no sample of play ever shown; writes out every weight and bias as a literal number};
\node[block, right=of llm] (net) {Frozen feed-forward network: fixed weights, one forward pass, no search, no further learning};
\node[block, right=of net] (ver) {Verification: exhaustive proof against an independent oracle, or statistical evaluation against an independent opponent};
\draw[arr] (spec) -- (llm);
\draw[arr] (llm) -- (net);
\draw[arr] (net) -- (ver);
\end{tikzpicture}%
}
\caption{The capability test that produces and judges every network in this
study. A language model receives a game's rules, the required tensor shapes, and
one solved example drawn from a different game, and must write out from
reasoning alone every entry of a feed-forward network's weights, so that the
resulting frozen forward pass plays the new game. The quantities that vary
across the paper are which game is posed, which model constructs, how much
code-execution tool access that model is given, and how the result is checked.
Each is a controlled variation of this single test.}
\label{fig:paradigm}
\end{figure}
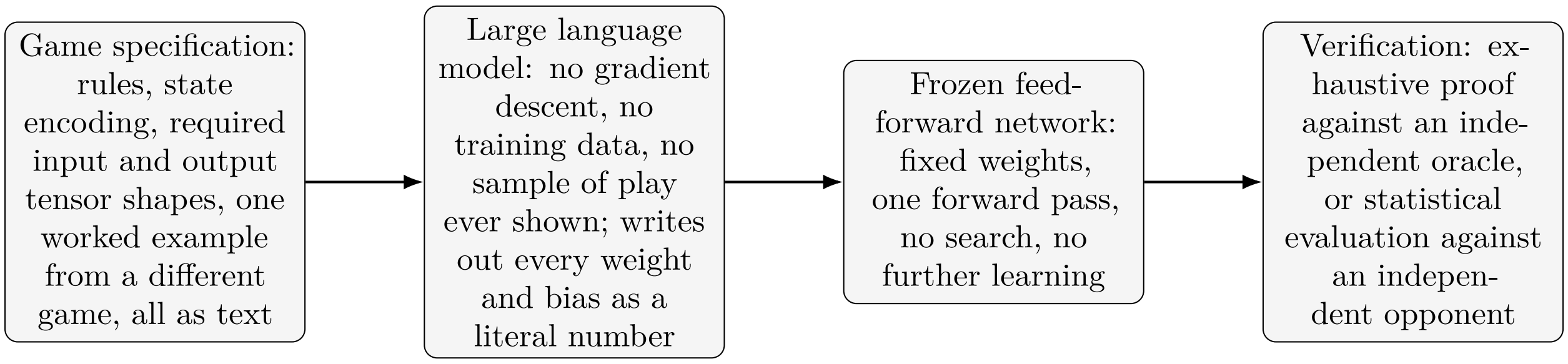

We measure the boundary on two-player games. Games are the right instrument for
this question because they come with exact oracles: for a game small enough to
enumerate, the game-theoretic value of every position can be computed
independently of the network under test, so correctness is decided rather than
estimated, and a failure cannot be attributed to an unlucky sample. Classical
game theory also supplies a difficulty gradient that is not merely a matter of
taste, since the games differ in whether an optimal strategy has a closed form,
in how large the reachable state space is, in the geometry of the action set,
and in how many moves ahead a competent player must look. We therefore build a
ladder of games that climbs that gradient one property at a time, ask a language
model to write every weight of a feed-forward network for each game from the
rules alone, and check each resulting network against an oracle written
separately from it. At every level of the ladder we repeat the exercise on a
second game with an unrelated mechanic, so that a finding about a level can be
distinguished from an accident of the one game used to probe it. Figure
\ref{fig:paradigm} shows the test that every one of the resulting networks is
produced and judged by.

The study contributes the following. First, a difficulty-graded suite of ten
hand-constructed game-playing networks, each written weight by weight with no
gradient descent and no training data, released with the independent oracles
that verify them. Second, an exhaustive correctness proof for six of the ten,
covering every reachable position of the game rather than a sample, at a scale
reaching 104\,399 positions, together with a complete-state-space check of the
remaining two small-board networks against independently computed chess endgame
tables. Third, the identification of a sharp and reproducible boundary: the
technique is exact where optimal play admits a local certificate that one
forward pass can check, and it collapses where competence requires composing
many local decisions into a plan, with the same two networks falling from 84.9
and 100 per cent at the one-ply tactic to 0.57 and 9.13 per cent in full
adversarial play. Fourth, a controlled attribution of that collapse to the
architecture of the constructed object, obtained by giving an opponent the same
hand-set evaluation and two plies of search, which wins every game. Fifth, six
further independent construction attempts spanning three providers and four
model sizes, which reproduce the boundary and show that it constrains what a
compact circuit can be derived to do rather than what the architecture can
represent. Sixth, two formal results that explain why the cost of an exact
construction is bounded for some games and grows with the state space for
others.

The paper is organized as follows. Section~\ref{sec:related} places the study
among its three neighbouring literatures: the weight-space programming work it
extends, the search-based and learning-based traditions of computer game
playing, and the combinatorial game theory from which the ladder's difficulty
claims are drawn. Section~\ref{sec:gametheory} states in game-theoretic terms
what the object under construction is and which two quantities the ladder
measures. Section~\ref{sec:method} describes the construction method: the gate
vocabulary a rulebook is compiled into, the two formal results that bound the
cost of doing so, the protocol under which the language model was asked to
construct, and the verification standard applied at each level.
Section~\ref{sec:experiments} reports the experiments, one subsection per level
of the ladder, followed by the cross-constructor replications, a prompting
intervention and an open-weight series. Section~\ref{sec:discussion} draws the
architectural reading the results support, and Section~\ref{sec:conclusion}
closes with limitations and future work.

\section{Related work}
\label{sec:related}

Three bodies of work meet in this study, and the contribution is easiest to see
once each is separated from the others. The first is the weight-space
programming literature itself, which supplies the technique and the name. The
second is computer game playing, which has pursued strong play in these same
games for seventy years and has done so, without exception, by one of two
routes: search at decision time, or weights obtained from data. The third is
combinatorial game theory, which supplies the exact results that make the
ladder's difficulty ordering a matter of record instead of judgement, and
retrograde analysis, which supplies the oracles the hardest rung is checked
against. We take them in that order.

\citet{weightspaceprogrammingblog} is the direct origin of the technique and of
the term used throughout this paper. Their language-model agent diagnoses,
repairs and writes neural network weights directly across eight experiments. Two
of those experiments are the closest to the present study. The first constructs
small game-playing networks from scratch for grid-world Atari-style games and
for three MinAtar environments, and reports that these outperform proximal
policy optimization baselines trained for five hundred million steps. The second
constructs a network for Go on a nine-by-nine board which, they report, wins
more than half its games against two smaller policy-gradient baselines given a
budget of zero additional forward passes at decision time, and beats an
equal-sized policy-gradient baseline once that budget is raised to fifty forward
passes. That second result is the closest point of contact with what we find
here. A budget of extra forward passes at decision time is exactly the resource
a single frozen forward pass does not have, and the fact that performance
improves as the budget grows from zero to fifty agrees with the architectural
reading we argue for in Section~\ref{sec:discussion}. Our companion study
\citep{garridomerchan2026tabular} applies the same practice to supervised
learning on structured data, where the target is a conditional expectation
rather than a strategy, and shares the definition of weight-space programming
restated in Section~\ref{sec:method}. Seen from outside, the practice is the
mirror image of mechanistic interpretability: that line of work starts from a
trained network and tries to recover the algorithm implicit in its weights,
while weight-space programming starts from an explicit algorithm and compiles it
into weights.

Computer game playing has produced strong players for every game on this ladder,
and the way it has done so is the relevant contrast. \citet{shannon1950chess}
set out the two options at the beginning: a program may apply a static
evaluation to the positions in front of it, or it may search ahead and apply the
evaluation at the leaves. The second option, formalized by the alpha-beta
analysis of \citet{knuth1975alphabeta}, carried the field through
\citet{campbell2002deepblue}, whose chess machine evaluated positions with a
hand-tuned function and searched on the order of a hundred million of them per
second. The networks constructed in this paper are the degenerate case of
Shannon's first option: the evaluation is hand-set, as in that tradition, and
the search budget is zero.

The second route replaces hand-set evaluation with weights learned from
experience. \citet{tesauro1995tdgammon} trained a backgammon evaluation by
temporal-difference learning on self-play and reached a standard of play
competitive with the strongest human players, which established that an
evaluation function need not be written by a person at all.
\citet{mnih2015dqn} extended the same idea to raw pixel input across Atari
games. \citet{silver2016alphago} combined a learned policy and value network
with Monte Carlo tree search to defeat a professional Go player;
\citet{silver2017alphagozero} removed the human games from the pipeline, and
\citet{silver2018alphazero} generalized the result to chess and shogi from
self-play alone. \citet{schrittwieser2020muzero} then removed the rules of the
game from the agent's inputs, learning a model of the dynamics along with the
policy and value. Beyond perfect information, \citet{brown2018libratus} and
\citet{vinyals2019alphastar} reach superhuman and grandmaster play in heads-up
poker and in StarCraft II. Every one of these systems obtains its weights from
data, whether supervised or generated by self-play, and every one of them except
the backgammon and Atari evaluators performs substantial search at decision
time. The present study removes both. The weights are derived from the rules by
reasoning, no position is ever evaluated during construction, and the policy
that results performs no search. The point of comparison is therefore the
resource these systems rely on, and the ladder measures what remains once that
resource is withdrawn.

The ladder's difficulty ordering is grounded in combinatorial game theory.
Nim's optimal strategy, under which a position is lost for the
player to move exactly when the bitwise exclusive-or of the pile sizes vanishes,
is the founding result of the field and is due to \citet{bouton1901nim}; the
Sprague-Grundy theory it opens onto \citep{sprague1935kampfspiele} is why we
excluded close relatives of Nim such as Kayles, which would repeat the
closed-form axis without adding a new one. Connect Four is a first-player win
achieved by opening in the centre column \citep{allis1988connectfour}, which is
the structural fact our standard-board network is checked against, since no
exact table is constructible by hand at that scale. Checkers is completely
solved \citep{schaeffer2007checkers}, but its solution is a 237-gigabyte
compressed database covering all 39 trillion legal positions, which places it on
the same side of the state-space axis as our own largest rungs without adding an
axis they do not already cover; we excluded it for that reason. The taxonomy of
\citet{herik2002gamessolved} supplies the vocabulary in which each rung's
verification standard is stated in Section~\ref{sec:standards}.

The hardest rung is checked against tables computed by retrograde analysis, the
technique that propagates exact game-theoretic values backward from checkmated
positions instead of searching forward from the current one. We implement it
from scratch for both chess endgames rather than importing an external
tablebase, so that the oracle shares no code and no data with the network it
judges. Before trusting it, we validate it twice against sources outside this
project: its position counts are compared with an independent public computation
of the same endgame by \citet{leslie-hurd-krk}, and its maximum distance to mate
with the value first established in Thomas Str\"ohlein's 1970 doctoral thesis,
as reported in the fiftieth-anniversary account of \citet{haworth2020strohlein}.
Section~\ref{sec:experiments} reports the outcome of both comparisons.

\section{The game-theoretic reading}
\label{sec:gametheory}

Section~\ref{sec:related} located this study between a technique and a
literature. This section says what the technique is being asked to produce,
using the vocabulary in which the literature states its results. We proceed in
four steps. We first identify the object under construction as a strategy in the
classical sense, then fix the notation and the optimality condition that every
later claim is stated against, then isolate the two quantities the ladder
measures, and finally state what the word "verified" means at each level of the
ladder.

Every game in this study is finite, two-player, zero-sum, of perfect
information, and free of chance moves. The classical theory of such games
settles, before a single weight is written down, what any player of one of them
is. \citet{zermelo1913} proved that in a game of this kind either one player can
force a win or both can force at least a draw, and that a forcing strategy never
needs more moves than the game has positions; \citet{schwalbe2001zermelo}
reconstruct the argument in modern notation and separate what Zermelo
established from what is commonly attributed to him. Backward induction on the
finite game tree turns that existence statement into a procedure.
\citet{kuhn1953extensive} supplies the formulation we use, in which a strategy is
a map from the positions at which a player moves to the actions available there,
and in which a game of perfect information always admits an optimal strategy that
is pure and reads only the current position. Each of the ten networks in this
paper is an attempt to realize one such strategy, or a controlled approximation
of one, as a single bounded-depth feed-forward circuit whose every coefficient is
fixed in advance from the rules of the game.

Making that attempt precise requires notation, which we fix once here. Let $S$ be
the finite set of positions at which the player being constructed is to move, let
$A(s)$ be the set of actions legal at a position $s \in S$, and let $T(s,a)$ be
the position reached from $s$ by playing $a \in A(s)$. Let $v(s) \in
\{-1,0,+1\}$ be the value of $s$ to the player to move under optimal play by both
sides, with $+1$ a forced win, $0$ a draw and $-1$ a forced loss. The game is
zero-sum and the two sides alternate, so the value of a position measured from
the side that has just moved is the negation of its value measured from the side
to move. A strategy $\sigma : S \to A$ with $\sigma(s) \in A(s)$ for every $s$ is
therefore optimal exactly when
\begin{align}
-\,v\bigl(T(s,\sigma(s))\bigr) \;=\; v(s) \qquad \text{for every } s \in S .
\label{eq:optimality}
\end{align}
The left-hand side is the value, measured from the perspective of the player at
$s$, of the position that $\sigma$ moves to. The right-hand side is what the
position was already worth to that player. The equation asks that playing
$\sigma$ give nothing away, which is the defining property of optimal play in
this class of games, and a strategy satisfying it everywhere is one side of a
subgame-perfect equilibrium whose existence Zermelo's theorem guarantees.
Equation~\eqref{eq:optimality} is the standard every exhaustive verification in
Section~\ref{sec:experiments} is run against.

Two questions about $\sigma$ are settled before the experiments begin, and
stating them is what isolates the question that is not. Whether $\sigma$ exists
is settled by Zermelo. Whether a feed-forward network can represent it is
settled by counting: for a finite $S$, a table of $|S|$ exact-match rows is
already such a network, and Section~\ref{sec:experiments} reports one built that
way for both chess endgames. The open question is the size of the circuit
needed. Fix the gate vocabulary of Section~\ref{sec:gates}, and define the
cost of a game to be the number of gates in the smallest network over that
vocabulary whose argmax satisfies \eqref{eq:optimality} at every position. This
is a circuit-size measure in the ordinary sense, relative to a fixed basis and a
fixed input encoding, and like every such measure it is a lower bound that no
experiment can observe directly; what a construction exhibits is an upper bound,
namely the size of the one circuit it managed to produce. We are therefore
careful throughout about which of the two we are reporting. The experiments
measure upper bounds, in a form that makes them comparable across games: after a
bounded set of hand-derived tactical conditions has been applied, the number of
entries the exact-match override table still has to carry is the residual
description length of equilibrium play under that particular tier vocabulary.
Proposition~\ref{prop:overridecost} is the one place where we convert this into
a statement about the minimum, and it does so only under a hypothesis strong
enough to rule out every tier vocabulary at once.

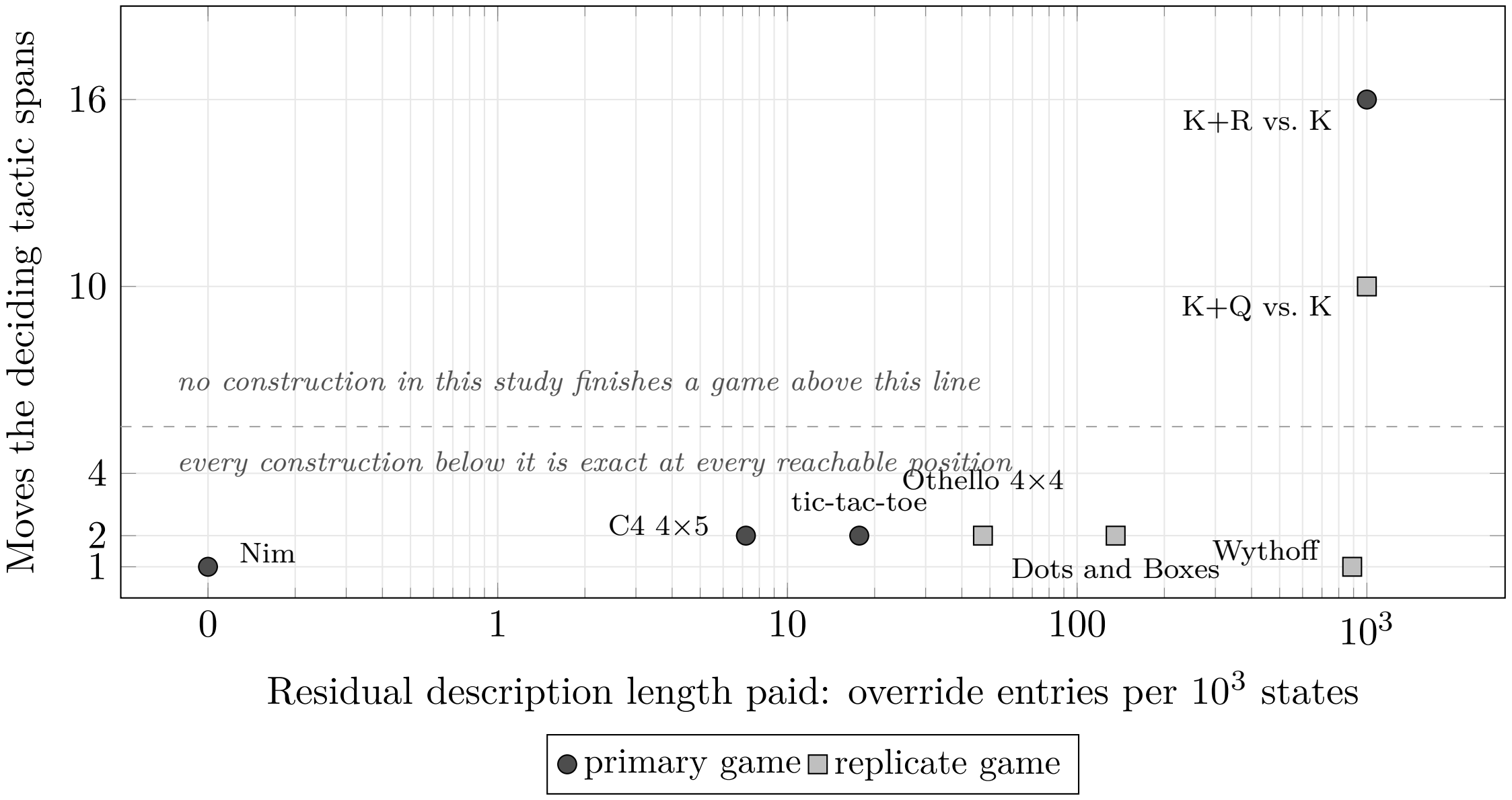
\begin{figure}[htbp]
\centering
\begin{tikzpicture}
\begin{axis}[
  width=0.92\linewidth,
  height=7.4cm,
  xlabel={Residual description length paid: override entries per $10^3$ states},
  ylabel={Moves the deciding tactic spans},
  xmode=log,
  xmin=0.05, xmax=3000,
  ymin=0, ymax=19,
  ytick={1,2,4,10,16},
  xtick={0.1,1,10,100,1000},
  xticklabels={$0$,$1$,$10$,$100$,$10^3$},
  grid=both,
  grid style={gray!18},
  every axis plot/.append style={mark size=2.6pt},
  legend style={at={(0.5,-0.23)}, anchor=north, legend columns=2, font=\small},
]
\addplot[only marks, mark=*, draw=black, fill=black!70] coordinates {
  (0.1,1) (17.7,2) (7.19,2) (1000,16)
};
\addlegendentry{primary game}
\addplot[only marks, mark=square*, draw=black, fill=black!25] coordinates {
  (890,1) (135.8,2) (47.3,2) (1000,10)
};
\addlegendentry{replicate game}
\node[font=\scriptsize, anchor=west] at (axis cs:0.115,1.45) {Nim};
\node[font=\scriptsize, anchor=east] at (axis cs:760,1.45) {Wythoff};
\node[font=\scriptsize, anchor=south] at (axis cs:17.7,2.35) {tic-tac-toe};
\node[font=\scriptsize, anchor=north] at (axis cs:135.8,1.72) {Dots and Boxes};
\node[font=\scriptsize, anchor=east] at (axis cs:6.0,2.3) {C4 $4{\times}5$};
\node[font=\scriptsize, anchor=south] at (axis cs:47.3,2.95) {Othello $4{\times}4$};
\node[font=\scriptsize, anchor=east] at (axis cs:850,15.3) {K+R vs.\ K};
\node[font=\scriptsize, anchor=east] at (axis cs:850,9.3) {K+Q vs.\ K};
\draw[dashed, gray!70] (axis cs:0.05,5.5) -- (axis cs:3000,5.5);
\node[font=\scriptsize\itshape, anchor=west, gray!60!black] at (axis cs:0.07,6.9)
  {no construction in this study finishes a game above this line};
\node[font=\scriptsize\itshape, anchor=west, gray!60!black] at (axis cs:0.07,4.3)
  {every construction below it is exact at every reachable position};
\end{axis}
\end{tikzpicture}
\caption{The eight exhaustively verified games on the plane of the two
quantities this section defines. The horizontal axis is the residual description
length our own constructions actually paid, measured as exact-match override
entries per thousand enumerable positions and plotted logarithmically; it is an
upper bound on the circuit size the game requires. The values are $0$ for Nim,
which needs no table at all and is drawn at the left-hand limit by convention,
$890$ for Wythoff, $17.7$ for tic-tac-toe, $135.8$ for Dots and Boxes, $7.2$ for
the reduced Connect Four board, $47.3$ for the reduced Othello board, and the
full $1000$ for the two chess endgames, whose only exact construction is a
tablebase carrying one entry per position. The vertical axis is the number of
moves the tactic that decides the position spans: one where a closed-form rule
settles the move outright, two where a two-move tactic decides it (the fork in
tic-tac-toe, the double-cross in Dots and Boxes, the trap directly above a
played cell in Connect Four, the corner-adjacency trap in Othello), and the
published maxima of sixteen and ten moves for the two endgames
\citep{haworth2020strohlein}. The dashed band separates the plane horizontally
and not diagonally, which is the finding Section~\ref{sec:discussion} develops:
the cost of description varies over four orders of magnitude without predicting
failure, while the span of the deciding tactic predicts it exactly. The two
standard-board games are absent because neither quantity is measurable for
them.}
\label{fig:axes}
\end{figure}

The second quantity the ladder measures is a property of the opponent rather
than of the game. \citet{shannon1950chess} separated a program that applies a
static evaluation to the positions in front of it from one that searches ahead
before applying that evaluation. The networks built here are the degenerate case
of the first kind: the forward pass reads a position and emits a move, and no
part of it recurses on its own output. The opponents they face in
Section~\ref{sec:experiments} are alpha-beta searches in the sense of
\citet{knuth1975alphabeta}, which perform a stated number of plies of backward
induction against a valuation of their own before committing to a move. The
comparison between them is a comparison between a player who best-responds to a
valuation and a player who iterates that reasoning a fixed number of times. That
is the same ladder of reasoning depths studied in experimental game theory under
the headings of iterated reasoning and cognitive hierarchies
\citep{stahl1995players,camerer2004cognitive}, with the difference that there the
depth is inferred from observed choices, while here it is imposed exactly by the
opponent's search depth. Figure~\ref{fig:axes} places all ten games of the
ladder on the plane spanned by these two quantities.

What the present data establish on the second axis needs stating carefully.
Every controlled comparison below fixes one opponent depth and varies the
construction, so the numbers show that a zero-lookahead construction loses at
the depths tested, without tracing the payoff as a function of depth. The
several depths appearing across the independent attempts of
Section~\ref{sec:replication} come from different hand-set valuations, so they
form a scatter rather than a curve, and Figure~\ref{fig:depthcurve} plots them
as such. Measuring that curve under a single fixed construction remains open.

The taxonomy of \citet{herik2002gamessolved} fixes what the word "verified"
amounts to at each level. A game is ultra-weakly solved when the value of its
initial position is known, weakly solved when a strategy achieving that value
from the initial position is known, and strongly solved when an optimal strategy
is known from every legal position. The six levels verified exhaustively here
are strongly solved by the construction and its independent oracle jointly,
since \eqref{eq:optimality} is checked at every element of $S$ with no sampling
anywhere. The two chess endgames are strongly solved by their retrograde
oracles, and the hand-built network is checked against that solution position by
position; what the construction fails to achieve at that level is the
compression of the solution into a circuit a person can derive. The two
standard-board games are solved in none of the three senses, which is why the
evidence there is statistical and why no override table exists to measure.

\section{Method}
\label{sec:method}

Section~\ref{sec:gametheory} described the target: a strategy satisfying
\eqref{eq:optimality}, realized as a bounded-depth circuit. This section
describes how such a circuit is produced and how the result is checked. We begin
with the problem the construction has to solve and the intuition that makes it
tractable, then state what counts as a weight-space-programmed network, then
give the gate vocabulary every construction in this paper is assembled from, then
bound the cost of that assembly with two formal results, then describe the
protocol under which a language model was asked to perform it, and finally state
the verification standard applied at each level of the ladder. A closing
subsection reports how the models actually proceed, which is an interpretation of
the constructions and is kept separate from the method for that reason.

The problem is one of compilation without a compiler. A game's rulebook is a
collection of discrete conditions: a move is legal when the destination square is
empty and no piece stands between it and the origin, a move wins when it
completes a line, a position is lost when a parity invariant vanishes. A
feed-forward network, by contrast, computes an alternating composition of affine
maps and rectifiers over real vectors. Nothing in the second object looks like
the first, and no procedure in standard practice converts one into the other: the
usual route is to abandon the rulebook, sample play, and let gradient descent
find coefficients that reproduce it. The question this paper studies is whether
the conversion can be done directly, by reasoning from the rules to the
coefficients, with no sample of play ever consulted.

The reason to want that conversion is that it changes what can be known about the
resulting network. A network fitted to samples is correct in a statistical sense
on the distribution it was fitted to, and its behaviour at any particular input
is an empirical question. A network compiled from the rules carries a derivation
for every coefficient, so its behaviour at every input is a mathematical
consequence of a stated construction, and can be checked exhaustively against an
independent implementation of the same rules. Six of this paper's ten networks
are checked that way at every reachable position, which is a standard no trained
network of comparable scope is ever held to.

The conversion is possible because a rectifier over binary inputs is a threshold
gate. If $x_1,\dots,x_k$ take values in $\{0,1\}$, then
$\mathrm{ReLU}(\sum_i x_i - (k-1))$ is one exactly when every $x_i$ is one, which
is a logical conjunction written as a single unit with integer weights. Boolean
connectives are what a rulebook is written in, so a layer of such units reads a
position and reports which rule-level conditions hold at it. What remains is to
convert a set of simultaneously true conditions into a single choice of move, and
that is achieved by scoring: each condition contributes to a candidate move's
score with a coefficient so much larger than everything beneath it that no
combination of lesser conditions can outweigh it, which makes the argmax of the
score respect a stated priority order without any branching in the forward pass.
The whole construction is therefore a bank of threshold gates feeding a weighted
sum, and both halves are things a model can write out coefficient by coefficient.

Before describing those pieces in detail we fix what the practice admits, since
the distinction from an ordinarily trained network is the entire point of the
exercise. The definition is stated for an arbitrary task, so that it applies
unchanged to the supervised setting of the companion study
\citep{garridomerchan2026tabular}.

\begin{Definition}
\label{def:wsp}
A feed-forward network $f_\theta$ is weight-space programmed for a task when
every entry of $\theta$ is fixed by a deterministic procedure, applied once
before $f_\theta$ is ever evaluated on an instance of the task, that takes as
input only a specification of the task and never takes as input a sample of the
network's own input-output behaviour. The procedure may be carried out by a
person, by a program, or by another neural network, and it may use arbitrary
non-differentiable computation, including exhaustive search over a finite space,
provided that the result of that computation enters $\theta$ as a constant and is
not invoked again at evaluation time.
\end{Definition}

Three consequences of Definition~\ref{def:wsp} are worth drawing out. The first
is what it forbids. Gradient descent is excluded, since a gradient of a loss is
computed from the network's own outputs. So are random search over parameters and
the write-then-test-then-revise loop a model with code execution would otherwise
fall into, for the same reason: each consults the artefact's behaviour to decide
the next value of $\theta$. The second is what it permits. A construction may
solve the game offline by exhaustive search and bake the answer into the weights,
which is what licenses the override tables used below, namely a set of reachable
positions where the single-pass tactical conditions do not by themselves match
optimal play, encoded as additional exact-match units. The forward pass still
performs no search; it evaluates fixed linear maps and rectifiers. Every network
in this study is accompanied by a test asserting that it holds no trainable
parameter, and all ten pass. The third consequence is the one that makes the
definition interesting here: the procedure is permitted to be another neural
network, and in every construction reported in this paper it is a large language
model, reading a specification in natural language and emitting literal numbers.

\subsection{The gate vocabulary}
\label{sec:gates}

\begin{figure}[htbp]
\centering
\begin{tikzpicture}[
  gin/.style={draw, circle, inner sep=1pt, minimum size=5.5mm, font=\scriptsize, fill=white},
  gop/.style={draw, rounded corners, minimum width=1.5cm, minimum height=6.5mm, font=\scriptsize, fill=gray!10},
  gout/.style={draw, circle, inner sep=1pt, minimum size=5.5mm, font=\scriptsize, fill=gray!25},
  garr/.style={-{Latex[length=1.6mm]}},
  gcap/.style={font=\scriptsize\itshape, anchor=west},
  gttl/.style={font=\small\bfseries, anchor=west},
]
\node[gttl] at (0,0) {Conjunction};
\node[gin] (a1) at (3.0,0.42) {$x_1$};
\node[gin] (a2) at (3.0,0.0) {$x_2$};
\node[gin] (a3) at (3.0,-0.42) {$x_3$};
\node[gop]  (as) at (5.1,0) {$\textstyle\sum\;-\,2$};
\node[gop]  (arl) at (7.1,0) {ReLU};
\node[gout] (ao) at (8.8,0) {$y$};
\draw[garr] (a1) -- (as); \draw[garr] (a2) -- (as); \draw[garr] (a3) -- (as);
\draw[garr] (as) -- (arl); \draw[garr] (arl) -- (ao);
\node[gcap] at (9.4,0) {$y=1$ iff all $x_i=1$};
\node[gttl] at (0,-1.7) {Disjunction};
\node[gin] (b1) at (3.0,-1.45) {$n_1$};
\node[gin] (b2) at (3.0,-1.95) {$n_2$};
\node[gop]  (bs) at (5.1,-1.7) {$\textstyle\sum\;-\,\tau$};
\node[gop]  (br) at (7.1,-1.7) {ReLU$\,\circ\,$ReLU};
\node[gout] (bo) at (8.8,-1.7) {$y$};
\draw[garr] (b1) -- (bs); \draw[garr] (b2) -- (bs);
\draw[garr] (bs) -- (br); \draw[garr] (br) -- (bo);
\node[gcap] at (9.4,-1.7) {clipped to $\{0,1\}$};
\node[gttl] at (0,-3.4) {Exact match};
\node[gin] (c1) at (3.0,-3.4) {$x$};
\node[gop]  (cs) at (5.1,-3.4) {$\cdot\,(2s{-}1)+c_s$};
\node[gop]  (cr) at (7.1,-3.4) {ReLU};
\node[gout] (co) at (8.8,-3.4) {$y$};
\draw[garr] (c1) -- (cs); \draw[garr] (cs) -- (cr); \draw[garr] (cr) -- (co);
\node[gcap] at (9.4,-3.4) {$y=1$ iff $x=s$};
\node[gttl] at (0,-5.1) {Tiered score};
\node[gin] (d1) at (3.0,-4.68) {$\phi_1$};
\node[gin] (d2) at (3.0,-5.1) {$\phi_2$};
\node[gin] (d3) at (3.0,-5.52) {$\phi_3$};
\node[gop]  (ds) at (5.6,-5.1) {$W_1\phi_1{+}W_2\phi_2{+}W_3\phi_3$};
\node[gout] (dz) at (8.8,-5.1) {$z_a$};
\draw[garr] (d1) -- (ds); \draw[garr] (d2) -- (ds); \draw[garr] (d3) -- (ds);
\draw[garr] (ds) -- (dz);
\node[gcap] at (9.4,-5.1) {$W_1\!\gg\!W_2\!\gg\!W_3$};
\end{tikzpicture}
\caption{The four gate primitives every network in this study is assembled from,
drawn as circuits. Conjunction thresholds the sum of $k$ binary indicators at
$k-1$, so the rectifier fires only when all of them hold. Disjunction clips a
nonnegative count into $\{0,1\}$, which takes two rectifier stages because one
cannot saturate. Exact match correlates the input against a stored state
encoded in $\pm 1$ and offsets by $c_s = |\{i : s_i = 0\}| - (n-1)$, firing on
that state alone. The tiered score multiplies each tactical indicator $\phi_t$
of a candidate action $a$ by a weight two or more orders of magnitude above the
next, so the argmax over $z_a$ follows the priority order of the tiers with no
branching anywhere in the forward pass. Every coefficient in all four is an
integer or a power of ten fixed by the construction, and none is fitted.}
\label{fig:gates}
\end{figure}
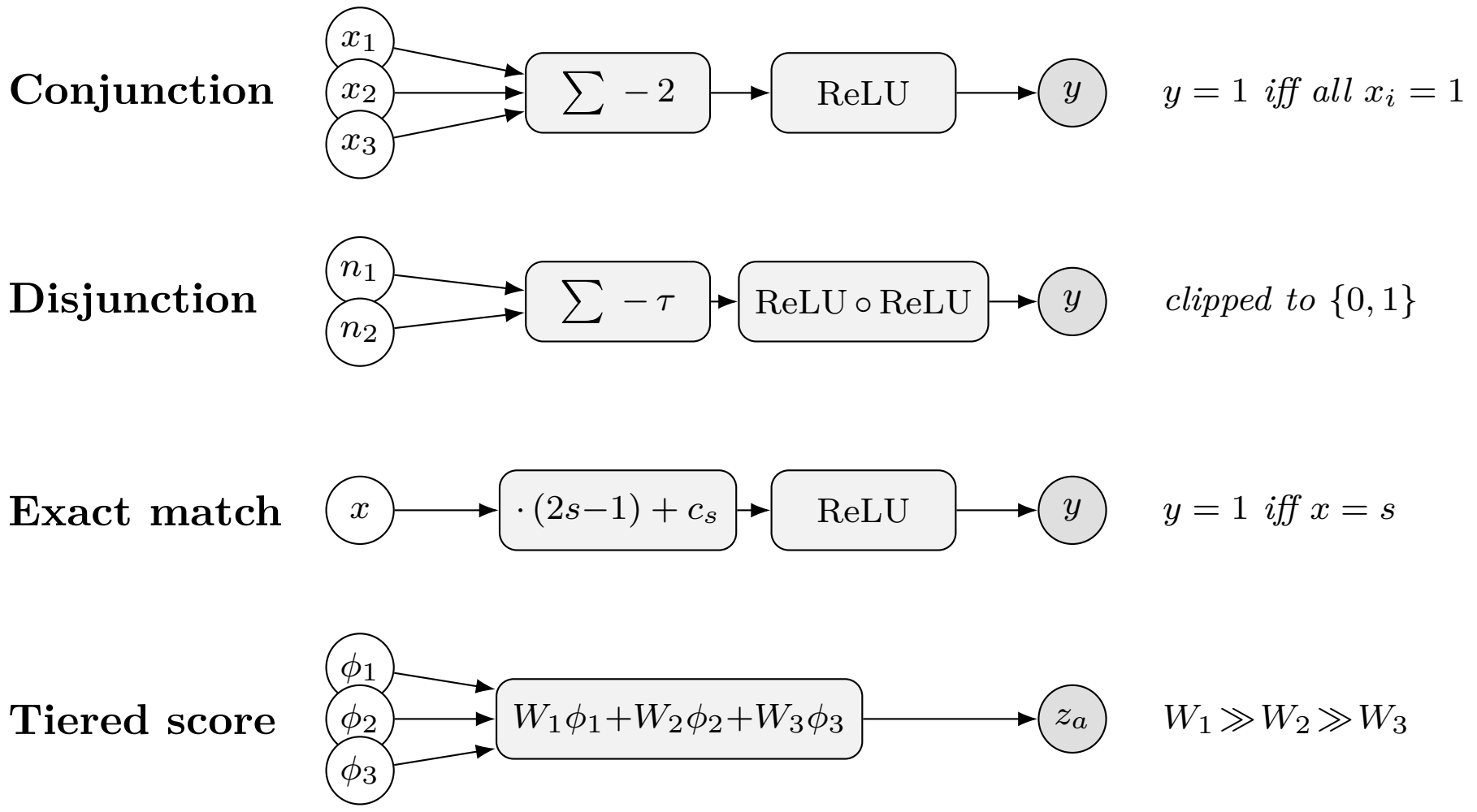

Four primitives account for every weight in the ladder, and the rest of the
construction is the choice of which conditions to express with them.
Figure~\ref{fig:gates} draws all four as circuits. Given binary indicators
$x_1,\dots,x_k$, the quantity $\mathrm{ReLU}(\sum_i x_i - (k-1))$ equals one if
every $x_i$ equals one and zero otherwise, which is a logical conjunction.
Disjunction is obtained by clipping: summing a set of nonnegative integer
signals, subtracting their threshold, applying a rectifier, subtracting the
result from one, applying a second rectifier, and subtracting that from one again
yields the indicator that at least one of the underlying events occurred, in two
stages rather than one because a single rectifier cannot saturate. For a stored
binary reference state $s \in \{0,1\}^n$ and a binary input $x$, the quantity
\begin{align}
\mathrm{ReLU}\!\left(x \cdot (2s-1) + \bigl|\{i : s_i = 0\}\bigr| - (n-1)\right)
\label{eq:exactmatch}
\end{align}
equals one exactly when $x$ agrees with $s$ in every coordinate. The dot product
against $2s-1$ rewards agreement with a one and penalizes disagreement, the
second term restores the count of coordinates where $s$ is zero, and the
threshold admits only the single input achieving the maximum. This is the
exact-match detector, used both for override tables and for the geometric
legality gates of the chess endgames. Finally, a tiered score over a finite set of
candidate actions assigns each tactical tier $t$ a scalar weight $W_t$ with $W_1
\gg W_2 \gg \cdots$, each at least two orders of magnitude above the next, so
that the largest possible total contribution from every tier below $t$ remains
smaller than a single unit of $W_t$. Summing $W_t$ times the tier-$t$ indicator
for each candidate action, adding a strictly smaller static positional bonus and
a strictly larger negative penalty on illegal actions, turns a hand-specified
priority ranking into one linear scoring function whose argmax always respects
that ranking.

One primitive that the four above do not supply is exclusive disjunction, which
Nim's optimal strategy requires and which no single rectifier computes, since the
parity function is not a threshold function. It is nonetheless available in two
stages, and we verify the identity by exhaustive case analysis over the four
binary input pairs before using it structurally.

\begin{Proposition}
\label{prop:xor}
For $a,b \in \{0,1\}$, $\mathrm{XOR}(a,b) = \mathrm{ReLU}(a+b) - 2\,\mathrm{ReLU}(a+b-1)$.
\end{Proposition}

\noindent When $a=b=0$ both rectifier terms vanish and the identity gives zero.
When exactly one of $a,b$ equals one, the first term equals one and the second
vanishes, giving one. When $a=b=1$ the first term equals two and the second
equals one, giving $2-2=0$. \qed

\noindent The identity matters beyond Nim because it shows what the vocabulary
buys. A function outside the threshold class is still reachable by composing two
rectifier stages. What bounds the construction is therefore its depth, since the
expressiveness of a single unit never binds. For Nim it means the bitwise
exclusive-or demanded by Bouton's theorem \citep{bouton1901nim} is two linear
layers and a rectifier, rather than a table.

\subsection{What an exact construction costs}
\label{sec:cost}

Nim's construction is cheap, and the reason is not special to Nim. It is worth
isolating, because the contrast with a game at the same level of the ladder that
fails to be cheap is the first quantitative finding of this study. The relevant
property is a classical one. For an impartial game under normal play, write
$\mathcal{G}(s)$ for the Grundy value of a position $s$, defined recursively as
the least non-negative integer that is not the Grundy value of any position
reachable from $s$ in one move. \citet{sprague1935kampfspiele} proved that
$\mathcal{G}(s)=0$ holds exactly at the positions lost for the player to move, so
a player who can recognize Grundy-zero positions one move ahead plays optimally.
Bouton's exclusive-or rule is the instance of this for Nim. The proposition below
states that whenever the Grundy-zero predicate itself compiles into the gate
vocabulary, the whole optimal strategy does, at a size that does not grow with the
number of positions.

\begin{Proposition}
\label{prop:grundy}
Let $G$ be an impartial game under normal play with finite position set $S$ and candidate action set $A$, positions encoded by an injection $\iota : S \to \{0,1\}^n$. For each $a \in A$ let $\lambda_a$ be the indicator that $a$ is legal at the encoded position, and let $\pi_a$ be the indicator that $a$ is legal and that the position $T(s,a)$ it reaches satisfies $\mathcal{G}(T(s,a)) = 0$. Suppose every $\lambda_a$ and every $\pi_a$ is computed on $\iota(S)$ by a ReLU circuit of depth at most $d$ and at most $w$ units, with $d$ and $w$ independent of $|S|$. Then the lexicographic tiered scorer of this section, taking $\pi_a$ as its top tier and $\lambda_a$ as its second, is a weight-space-programmed network of depth $d + O(1)$ and $O(|A|\,w)$ parameters whose argmax satisfies \eqref{eq:optimality} at every position of $S$.
\end{Proposition}

\noindent The argument has two cases and a count. Suppose first that $\mathcal{G}(s) \neq 0$. By the Sprague-Grundy property some legal action reaches a position of Grundy value zero, so at least one $\pi_a$ equals one, and the two-case dominance of the tiered scorer makes the argmax an action with $\pi_a = 1$. That action reaches a position lost for the opponent, whose value to the opponent is therefore $-1$, so the left-hand side of \eqref{eq:optimality} equals $+1$, which is $v(s)$ because a position of non-zero Grundy value is won for the player to move. Suppose instead that $\mathcal{G}(s) = 0$. Then no legal action reaches Grundy value zero, every $\pi_a$ vanishes, the argmax falls to the second tier and returns some legal action, and every such action reaches a position won for the opponent, so the left-hand side of \eqref{eq:optimality} equals $-1$, which is again $v(s)$. For the count, the network holds one copy of each $\pi_a$ and $\lambda_a$ circuit, none of whose widths depends on $|S|$, together with the single linear layer that applies the tier weights, which gives $O(|A|\,w)$ parameters and depth $d$ plus the constant depth of the scorer. \qed

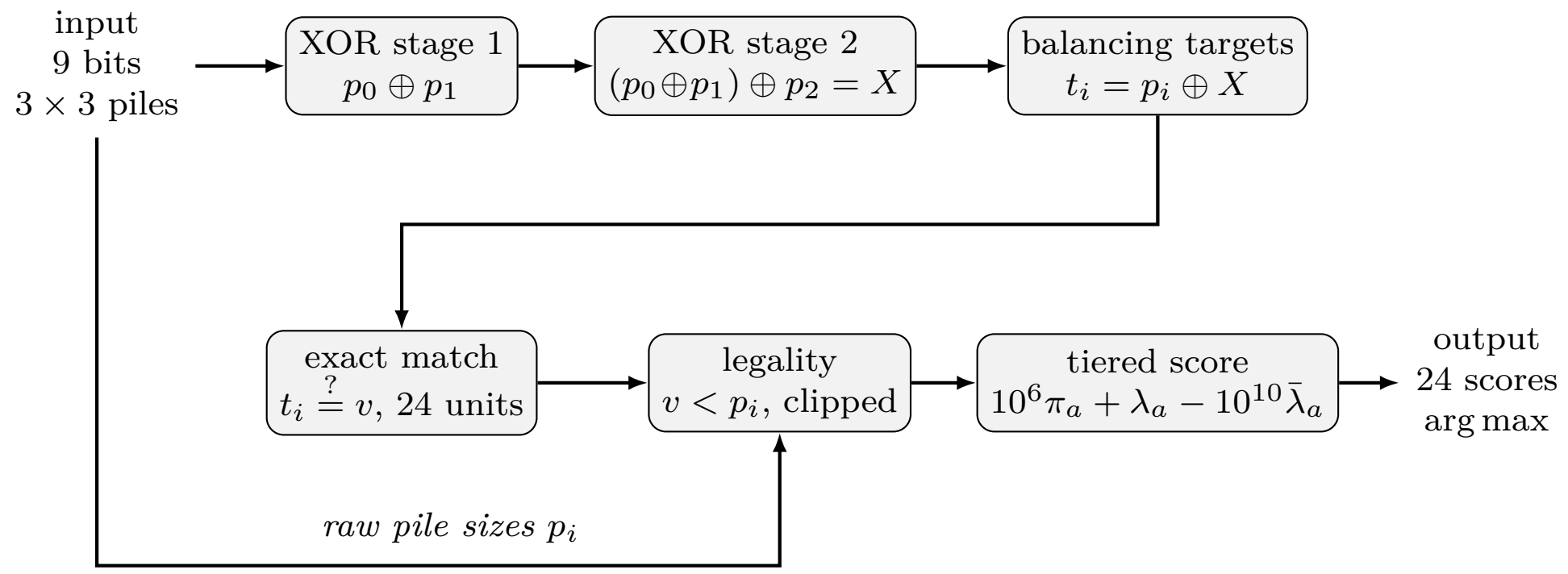
\begin{figure}[htbp]
\centering
\begin{tikzpicture}[
  nlay/.style={draw, rounded corners, minimum height=8mm, font=\scriptsize, fill=gray!10, align=center, inner sep=3pt},
  ndat/.style={font=\scriptsize, align=center},
  narr/.style={-{Latex[length=1.6mm]}, thick},
]
\node[ndat] (in) at (0,0) {input\\$9$ bits\\$3\times3$ piles};
\node[nlay] (x1) at (2.5,0) {$\mathrm{XOR}$ stage 1\\$p_0 \oplus p_1$};
\node[nlay] (x2) at (5.4,0) {$\mathrm{XOR}$ stage 2\\$(p_0\!\oplus\! p_1)\oplus p_2 = X$};
\node[nlay] (tg) at (8.7,0) {balancing targets\\$t_i = p_i \oplus X$};
\node[nlay] (mt) at (2.5,-2.6) {exact match\\$t_i \stackrel{?}{=} v$, $24$ units};
\node[nlay] (lg) at (5.6,-2.6) {legality\\$v < p_i$, clipped};
\node[nlay] (sc) at (8.7,-2.6) {tiered score\\$10^{6}\pi_a + \lambda_a - 10^{10}\bar\lambda_a$};
\node[ndat] (outp) at (11.4,-2.6) {output\\$24$ scores\\$\arg\max$};
\draw[narr] (in) -- (x1); \draw[narr] (x1) -- (x2); \draw[narr] (x2) -- (tg);
\draw[narr] (tg.south) -- (8.7,-1.3) -- (2.5,-1.3) -- (mt.north);
\draw[narr] (mt) -- (lg); \draw[narr] (lg) -- (sc); \draw[narr] (sc) -- (outp);
\draw[narr] (in.south) -- (0,-4.1) -- (5.6,-4.1) -- (lg.south);
\node[font=\scriptsize\itshape, anchor=south] at (2.9,-4.05) {raw pile sizes $p_i$};
\end{tikzpicture}
\caption{The Nim network, which is the construction
Proposition~\ref{prop:grundy} describes, drawn end to end. Two applications of
Proposition~\ref{prop:xor} compute the exclusive-or $X$ of the three pile sizes,
a third gives each pile its balancing target $t_i = p_i \oplus X$, and one
exact-match unit per candidate action asks whether that action moves pile $i$ to
the value $t_i$ demands, which is exactly the Grundy-zero indicator $\pi_a$. The
legality indicator $\lambda_a$ is a clipped comparison. The tiered score places
$\pi_a$ above $\lambda_a$ and both above a dominating penalty on illegal
actions, so the argmax balances the position whenever a balancing move exists and
plays legally otherwise. No entry of the network depends on the number of
positions: the width is set by the twenty-four candidate actions, which is the
content of the proposition's $O(|A|w)$ bound. The whole network holds $4323$
parameters, none of them trainable.}
\label{fig:nimcircuit}
\end{figure}

\noindent The hypothesis is satisfied whenever the Grundy-zero predicate has a
bounded circuit and each action's effect on the encoding is a fixed linear map,
since prepending a linear map to a rectifier circuit absorbs into its first
layer. Nim satisfies it through Proposition~\ref{prop:xor}, which yields, for
each pile and each target value, whether moving that pile to that value balances
the position. Figure~\ref{fig:nimcircuit} shows the resulting network in full.
Its size grows only through the one score per candidate action that any scorer of
this form must emit, so it is independent of the $|S|$ positions those actions are
scored at, which is what the proposition's parameter count asserts.

The hypothesis of Proposition~\ref{prop:grundy} is not idle, and the ladder
contains a game at the same level that fails it. Wythoff's game
\citep{wythoff1907nim} also has a closed-form optimal strategy, but one that
exposes a distinction the exclusive-or identity hides, since not every closed
form is equally cheap to compile. A position is losing for the player to move
exactly when its two pile sizes equal $\lfloor n\varphi\rfloor$ and $\lfloor
n\varphi^2\rfloor$ for some non-negative integer $n$, where $\varphi$ is the
golden ratio. Unlike the exclusive-or, $\lfloor n\varphi\rfloor$ has no
finite-size rectifier circuit computing it exactly for every $n$, because
$\varphi$ is irrational while a bounded-depth network of bounded-precision linear
maps represents only a rational function of its inputs. The only construction
available within this vocabulary is the exact-match lookup of
\eqref{eq:exactmatch}, applied here to the entire bounded state space rather than
to a handful of exceptions: every losing position within the bounded pile range
is precomputed once, offline, and baked in as its own unit. The qualitative claim
that a closed-form optimal strategy compiles into verifiable hand-set weights
survives this replicate. The quantitative cost does not transfer, since Nim's
strategy costs a number of gates that grows only with the number of candidate
actions while Wythoff's costs one gate per reachable losing position.

That asymmetry recurs at every later level of the ladder, wherever a bounded set
of tactical tiers is combined with an exact-match override table, so it is worth
stating once in general rather than re-deriving informally each time. The
following proposition says that the table's size is exactly the number of
positions where the tiers disagree with the target, and that this number is
forced to grow with the state space whenever the target is not nearly constant on
the classes of positions the tiers cannot tell apart.

\begin{Proposition}
\label{prop:overridecost}
Fix a tiered-plus-override scorer of the kind defined above: $k$ Boolean tier predicates $\phi_1,\dots,\phi_k$, each a fixed function of a state $s$ and a candidate action $a$ from a bounded, $|S|$-independent action set, combined by the lexicographic rule of this section, plus an override table whose bonus for a stored state dominates every tier's maximum combined contribution. Let $f_0 : S \to A$ be the action this scorer selects at each state before any override is consulted, and let $f : S \to A$ be a target action function the scorer is required to match exactly on every state of a finite state space $S$. The scorer computes $f$ everywhere if and only if its override table contains exactly the disagreement set $D := \{s \in S : f_0(s) \neq f(s)\}$, so the smallest table that works has size $|D|$. Since each $\phi_t$ is binary, the tuple $(\phi_1(s,a),\dots,\phi_k(s,a))_{a \in A(s)}$ takes at most $2^{k|A|}$ values, a bound independent of $|S|$, and $f_0$ is constant on every class of states sharing a tuple value; consequently $|D|$ can be smaller than $|S|$ only to the extent that $f$ itself is close to constant on a bounded number of such classes. In particular, if $f$ provably disagrees with every constant assignment on all but finitely many states of every such class, for every choice of $k$ tiers drawn from the available gate vocabulary, then $|D|$, and hence the minimum override table size, grows without bound as $|S|$ grows, regardless of how the $k$ tiers are chosen.
\end{Proposition}

\noindent The two-case dominance built into the scorer means the argmax at a stored state is always the table's prescribed action and the argmax at every other state is $f_0(s)$, which gives the stated equality; the bound on the number of tier-value classes is immediate since $k$ and $|A|$ are fixed while $|S|$ is not. \qed

\noindent Wythoff's game satisfies the hypothesis of the final clause exactly.
The sequence $\lfloor n\varphi \rfloor$ has irrational slope, so no fixed-size
arithmetic-threshold tier, meaning a comparison of $n$ against a constant built
from the bounded-precision linear maps this vocabulary provides, can agree with
Beatty-sequence membership beyond a bounded range of $n$ without being re-tuned
as that range grows. The override table is therefore forced to scale with the
pile bound itself, which is the cost measured in Section~\ref{sec:rung0}. We do
not claim the same forced scaling for the other tables this paper reports.
Those costs are measured rather than derived from an impossibility argument, and
a larger or cleverer tier vocabulary might shrink them; the proposition explains
why an exact construction can be forced to pay for the full state space, and does
not assert that any particular table reported here is minimal.

\subsection{The construction protocol}
\label{sec:protocol}

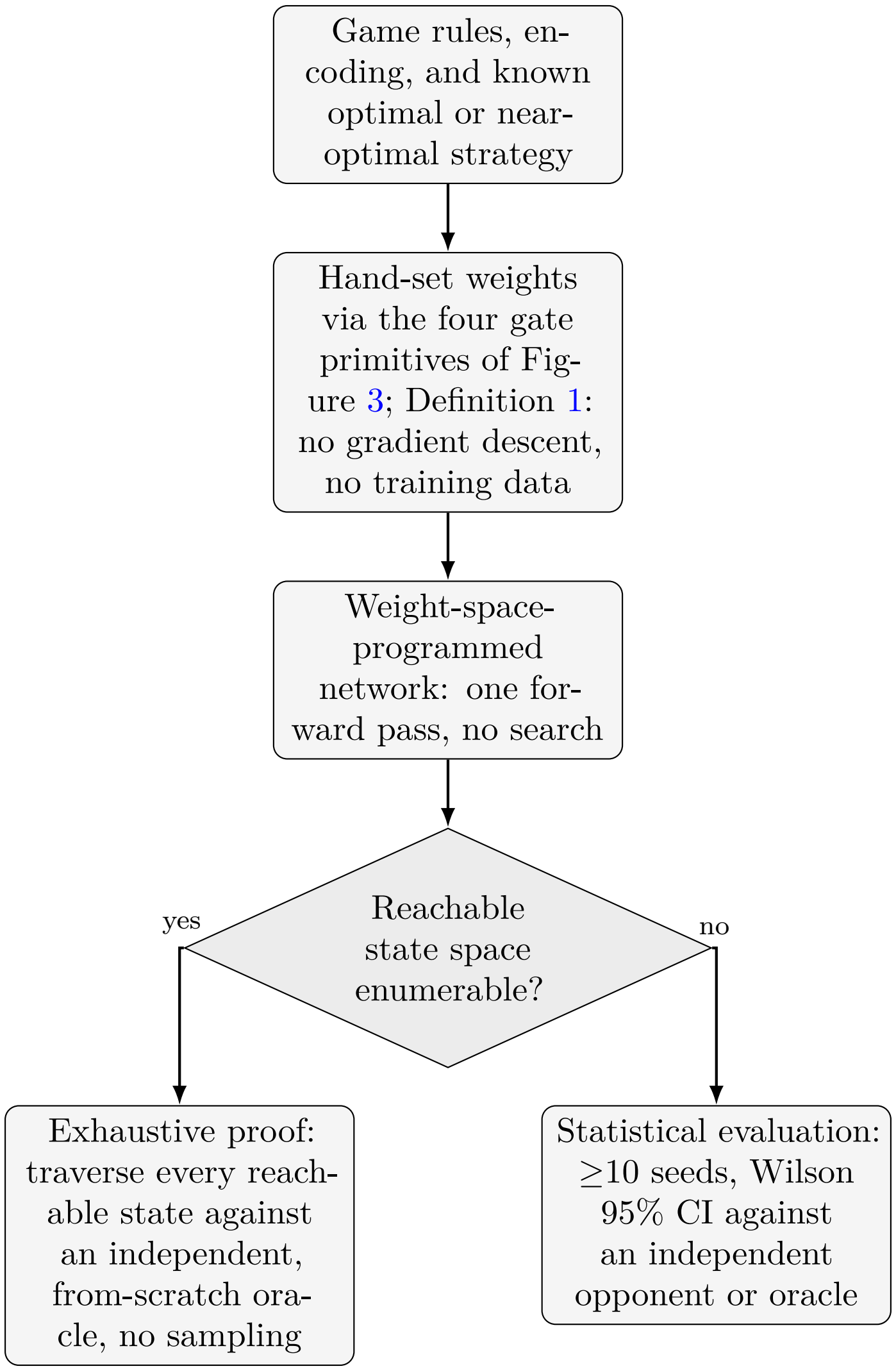
\begin{figure}[htbp]
\centering
\begin{tikzpicture}[
  node distance=0.7cm and 1.2cm,
  block/.style={draw, rounded corners, align=center, fill=gray!8, minimum width=3.6cm, minimum height=1.1cm, font=\small, text width=3.3cm},
  decision/.style={draw, diamond, aspect=2.2, align=center, fill=gray!15, font=\small, text width=2.6cm, inner sep=2pt},
  arr/.style={-{Latex[length=2mm]}, thick},
]
\node[block] (game) {Game rules, encoding, and known optimal or near-optimal strategy};
\node[block, below=of game] (weights) {Hand-set weights via the four gate primitives of Figure~\ref{fig:gates}; Definition~\ref{def:wsp}: no gradient descent, no training data};
\node[block, below=of weights] (net) {Weight-space-programmed network: one forward pass, no search};
\node[decision, below=of net] (dec) {Reachable state space enumerable?};
\node[block, below left=1.0cm and -0.4cm of dec] (exact) {Exhaustive proof: traverse every reachable state against an independent, from-scratch oracle, no sampling};
\node[block, below right=1.0cm and -0.4cm of dec] (stat) {Statistical evaluation: $\geq$10 seeds, Wilson 95\% CI against an independent opponent or oracle};

\draw[arr] (game) -- (weights);
\draw[arr] (weights) -- (net);
\draw[arr] (net) -- (dec);
\draw[arr] (dec) -| node[above, pos=0.25, font=\scriptsize] {yes} (exact);
\draw[arr] (dec) -| node[above, pos=0.25, font=\scriptsize] {no} (stat);
\end{tikzpicture}
\caption{The construction-and-verification pipeline applied to every network in
this study. A game's rules and, where one exists, its closed-form strategy are
compiled by hand into weights using the four primitives of
Figure~\ref{fig:gates}, producing a single forward pass with no gradient descent
and no search anywhere in the construction. The resulting network is then checked
exhaustively against an independent oracle wherever the reachable state space is
small enough to enumerate, and statistically otherwise.}
\label{fig:pipeline}
\end{figure}

What the model is given, and what it is denied, determines what the resulting
measurement means. Every construction in this study starts from one text
document containing the rules of the target game, the encoding of a position as a
binary vector, the required input and output tensor shapes, and one worked
example of the technique applied to a different and simpler game. The worked
example is the tic-tac-toe network, which existed in this project before the
present study and which is reused for that purpose; it is the only place a model
is shown what a finished construction looks like, and it is drawn from a game at
the second level of the ladder, so no model ever sees a solution to the game it
is asked to solve. The gate constructions specific to this paper's own solutions,
the exclusive-or identity of Proposition~\ref{prop:xor} and the geometric
legality gadgets of the chess endgames, are withheld from the brief, because
deriving them is the skill under test. Figure~\ref{fig:pipeline} shows the
pipeline that this brief enters and that a verified network leaves.

The amount of tooling the constructing model is allowed is the one condition we
vary deliberately, and we separate the two settings so that results obtained
under them are never averaged together. Track A gives the model full code
execution, iterative testing against its own test suite, and exhaustive-search
tooling during construction. Every network of the main ladder is built this way,
and every number in Sections~\ref{sec:rung0} to~\ref{sec:rung3} is produced under
it. Track B withdraws all of that. It administers one self-contained prompt per
game to a model with no tool access and a single attempt, containing a
plain-language statement of the method, the tic-tac-toe worked example, and the
specification of the new game, and it grades every submission with the same
independent oracles Track A is checked against. Track B covers the five primary
games rather than their replicates, since the replicates exist to validate the
ladder's findings and not to extend the cross-model protocol. Mixing an iterated,
tool-assisted result with a single cold attempt would not be a controlled
comparison, which is why the two tracks are reported separately throughout.

\subsection{Verification standards}
\label{sec:standards}

The standard against which a construction is judged changes deliberately as the
ladder climbs, and the change tracks what the state space permits rather than
what is convenient. Three standards are used. The first applies wherever the
reachable state graph can be traversed completely: Nim, Wythoff's game,
tic-tac-toe, Dots and Boxes, and the two reduced boards. For every non-terminal
state, under every possible sequence of legal replies by an opponent who may play
any legal move at each of its turns, the outcome the network achieves is checked
against the oracle's exact game-theoretic value for that state, with no sampling
anywhere. The second applies to the two chess endgames, where traversing complete
game trees is infeasible although enumerating positions is not. For every legal
position with the network to move, we check whether its chosen move strictly
decreases the independently computed distance-to-mate potential, which is
exhaustive over positions rather than over game trees and equally free of
sampling, and we separately simulate a large fixed sample of complete games
against an opponent that selects, at each of its turns, the reply worst for the
network, to measure whether per-move progress composes into finished games within
a generous move cap. The third applies once the state space is not enumerable at
all, at standard Connect Four and standard Othello: fixed random seeds, at least
ten of them, a stated number of games per seed and per condition, and Wilson
score 95 per cent confidence intervals on the resulting win rate. Every oracle is
written from scratch for this study and shares no code with the network it
judges.

\subsection{How the models construct: an interpretation}
\label{sec:interpretation}

\begin{figure}[htbp]
\centering
\resizebox{\textwidth}{!}{%
\begin{tikzpicture}[
  tstep/.style={draw, rounded corners, fill=gray!8, font=\scriptsize, align=center, text width=2.45cm, minimum height=1.3cm, inner sep=3pt},
  tnote/.style={font=\scriptsize\itshape, align=center, text width=2.6cm, gray!50!black},
  tarr/.style={-{Latex[length=1.8mm]}, thick},
]
\node[tstep] (s1) at (0,0) {1. Name the rules as Boolean conditions on a position};
\node[tstep] (s2) at (3.1,0) {2. Fix the encoding and the candidate action set};
\node[tstep] (s3) at (6.2,0) {3. Rank the conditions into tactical tiers};
\node[tstep] (s4) at (9.3,0) {4. Write each condition with a gate primitive};
\node[tstep] (s5) at (12.4,0) {5. Check penalty and tier margins arithmetically};
\draw[tarr] (s1) -- (s2); \draw[tarr] (s2) -- (s3); \draw[tarr] (s3) -- (s4); \draw[tarr] (s4) -- (s5);
\node[tnote] at (0,-1.55) {a condition not named here cannot be expressed later};
\node[tnote] at (3.1,-1.55) {one-hot per cell or per piece in every attempt};
\node[tnote] at (6.2,-1.55) {where the double-cross forces the override above the tactic};
\node[tnote] at (9.3,-1.55) {rediscovered independently by every attempt};
\node[tnote] at (12.4,-1.55) {the step most often skipped, and the hardest failure to see};
\node[tstep, fill=gray!18] (s6) at (12.4,2.0) {6. With tools only: solve offline, add one override per disagreement};
\draw[tarr, dashed] (s5.north) -- (s6.south);
\node[font=\scriptsize\itshape, gray!50!black, anchor=east] at (11.0,2.0)
  {the step that turns a heuristic into an exact construction};
\end{tikzpicture}%
}
\caption{The construction sequence observed across the seven attempts reported in
Section~\ref{sec:replication}, with what each step determines. The first four
steps occur in every attempt, with or without tool access, and the fifth is the
one most often omitted. The sixth is available only under Track A, where code
execution permits an offline exhaustive solve, and it is the step that converts a
tiered heuristic into a construction exact at every reachable position. This
figure summarizes transcripts and submitted modules; it is an interpretation of
how the constructions were reached and not a measurement of any network.}
\label{fig:trace}
\end{figure}
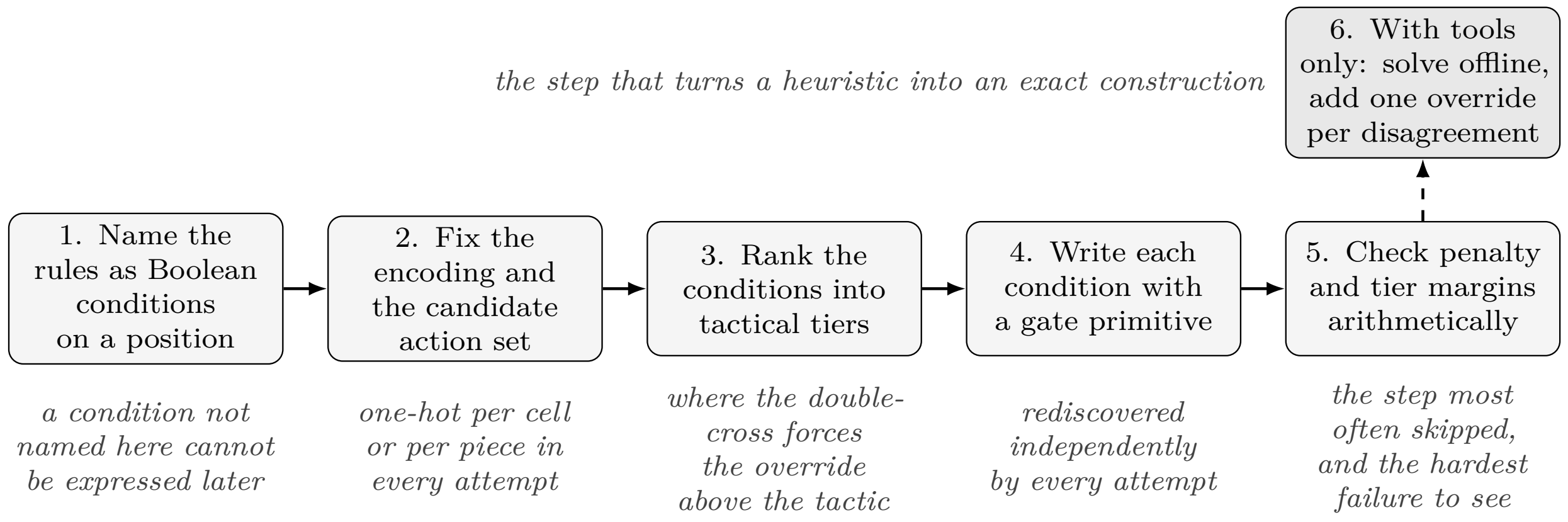

Everything above describes what was done. This subsection reports what the
construction process looked like from outside, which is an interpretation of the
transcripts and the resulting modules rather than a measurement, and we label it
as such so that no later claim rests on it. Across the seven construction
attempts reported in Section~\ref{sec:replication}, a common sequence recurs, and
Figure~\ref{fig:trace} sets it out. A model first restates the rules as a list of
Boolean conditions on a position, which is the step that determines everything
downstream, since a condition it fails to name cannot be expressed later. It then
chooses an encoding, almost always one-hot per cell or per piece, and fixes the
action set. It then ranks the conditions into tiers by tactical priority, and only
at that point does it reach for the gate vocabulary, writing each condition as a
conjunction, a clipped disjunction or an exact-match unit, and assigning tier
weights separated by powers of ten. The final step is arithmetic bookkeeping:
checking that the illegal-action penalty exceeds the largest attainable legal
score, and that each tier's weight exceeds the total of everything beneath it.
Where a model has code execution it closes with an exhaustive comparison against
its own oracle and adds an override entry for every disagreement, which is the
step that converts a heuristic into an exact construction.

Two features of that sequence are worth recording because they recur across
providers. The first is that the gate vocabulary is not given to the models and is
nonetheless rediscovered: every attempt that produced a literal circuit arrived at
the conjunction-as-threshold identity independently, and three of them arrived
independently at the conclusion that Wythoff's game admits no finite rectifier
circuit and must be tabulated. The second is that the step most often skipped is
the arithmetic bookkeeping, and the failures it produces are the ones that look
most like success, since a network with an insufficient illegal-action penalty
plays legally almost everywhere and illegally in the rare position where a tier
stacks high enough to overcome it. Section~\ref{sec:prompting} tests whether
demanding that step explicitly in the prompt improves the outcome.

\section{Experiments}
\label{sec:experiments}

The method of Section~\ref{sec:method} is applied here to ten games arranged as
a ladder, and this section reports what each level of that ladder yields. We
first explain how the ladder is designed, since the design is what allows a
failure to be attributed to a property of the games rather than to an
unstructured increase in difficulty. The five subsections that follow take the
levels in order, each presenting its games, its results and the interpretation
those results support. Three further subsections then vary what the earlier ones
hold fixed: the model doing the constructing, the prompt it is given, and the
scale of the open-weight models the protocol can reach.

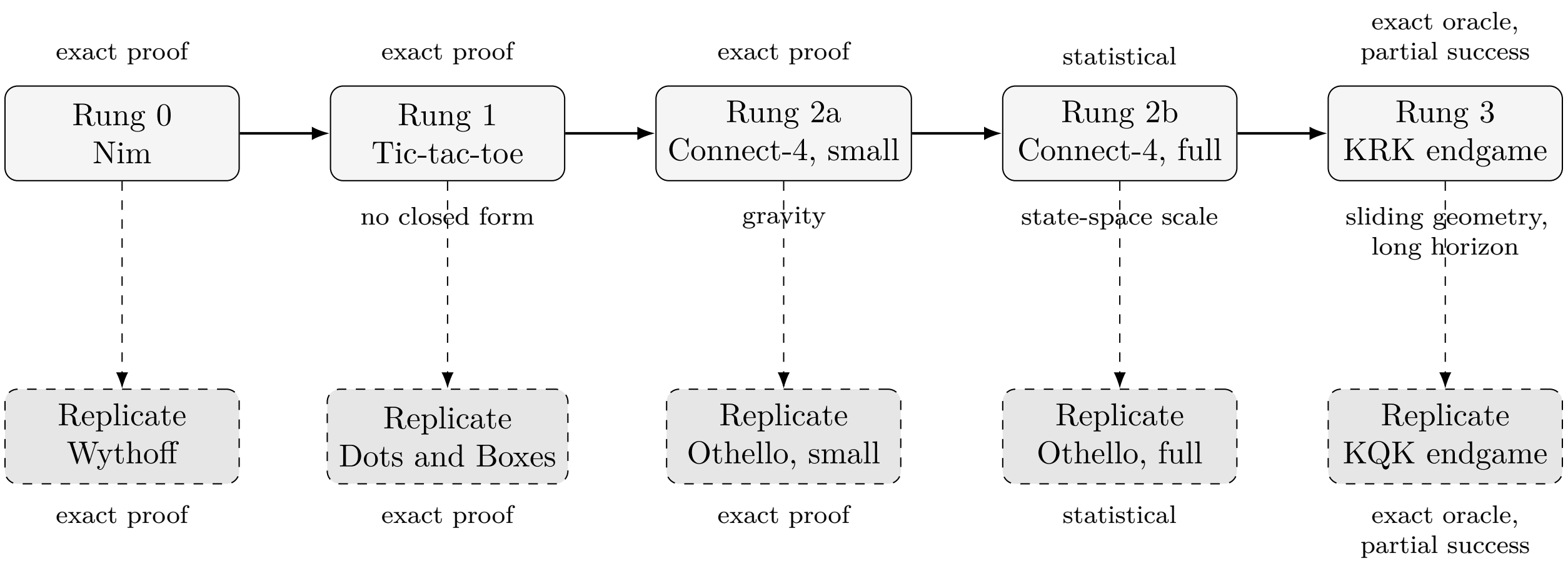
\begin{figure}[htbp]
\centering
\resizebox{\textwidth}{!}{%
\begin{tikzpicture}[
  node distance=0.9cm and 1.0cm,
  block/.style={draw, rounded corners, align=center, fill=gray!8, minimum width=2.6cm, minimum height=1.05cm, font=\small},
  rep/.style={draw, rounded corners, align=center, fill=gray!20, minimum width=2.6cm, minimum height=1.05cm, font=\small, dashed},
  arr/.style={-{Latex[length=2mm]}, thick},
  rarr/.style={-{Latex[length=1.8mm]}, dashed},
  lbl/.style={font=\scriptsize, align=center, text width=2.4cm}
]
\node[block] (n0) {Rung 0\\Nim};
\node[block, right=of n0] (n1) {Rung 1\\Tic-tac-toe};
\node[block, right=of n1] (n2a) {Rung 2a\\Connect-4, small};
\node[block, right=of n2a] (n2b) {Rung 2b\\Connect-4, full};
\node[block, right=of n2b] (n3) {Rung 3\\KRK endgame};

\draw[arr] (n0) -- (n1);
\draw[arr] (n1) -- (n2a);
\draw[arr] (n2a) -- (n2b);
\draw[arr] (n2b) -- (n3);

\node[lbl, below=0.15cm of n1] {no closed form};
\node[lbl, below=0.15cm of n2a] {gravity};
\node[lbl, below=0.15cm of n2b] {state-space scale};
\node[lbl, below=0.15cm of n3] {sliding geometry,\\long horizon};

\node[font=\scriptsize, align=center, above=0.1cm of n0] {exact proof};
\node[font=\scriptsize, align=center, above=0.1cm of n1] {exact proof};
\node[font=\scriptsize, align=center, above=0.1cm of n2a] {exact proof};
\node[font=\scriptsize, align=center, above=0.1cm of n2b] {statistical};
\node[font=\scriptsize, align=center, above=0.1cm of n3] {exact oracle,\\partial success};

\node[rep, below=2.3cm of n0] (r0) {Replicate\\Wythoff};
\node[rep, below=2.3cm of n1] (r1) {Replicate\\Dots and Boxes};
\node[rep, below=2.3cm of n2a] (r2a) {Replicate\\Othello, small};
\node[rep, below=2.3cm of n2b] (r2b) {Replicate\\Othello, full};
\node[rep, below=2.3cm of n3] (r3) {Replicate\\KQK endgame};

\draw[rarr] (n0) -- (r0);
\draw[rarr] (n1) -- (r1);
\draw[rarr] (n2a) -- (r2a);
\draw[rarr] (n2b) -- (r2b);
\draw[rarr] (n3) -- (r3);

\node[font=\scriptsize, align=center, below=0.1cm of r0] {exact proof};
\node[font=\scriptsize, align=center, below=0.1cm of r1] {exact proof};
\node[font=\scriptsize, align=center, below=0.1cm of r2a] {exact proof};
\node[font=\scriptsize, align=center, below=0.1cm of r2b] {statistical};
\node[font=\scriptsize, align=center, below=0.1cm of r3] {exact oracle,\\partial success};
\end{tikzpicture}%
}
\caption{The ladder in the top row, with solid arrows, and its replicates in the
bottom row, with dashed connectors. A solid arrow changes exactly one of the four
difficulty axes relative to the level before it; a dashed connector changes the
game mechanic while holding every axis fixed. We use the word rung throughout for
one position on this ladder: a fixed combination of values on the four axes,
instantiated by one primary game and one structurally unrelated replicate, built
and verified independently of each other. Rungs 2a and 2b hold every axis fixed
except state-space size, which is why they share a number.}
\label{fig:ladder}
\end{figure}

A ladder of games is only informative if a step along it changes one thing. We
therefore vary, one at a time wherever the structure of the axis allows it, four
properties that plausibly govern whether a game's optimal or near-optimal policy
can be compiled by hand into a bounded-depth circuit. The first is the size of
the reachable state space, which decides whether correctness can be proved
exhaustively or only estimated. The second is the existence of a closed-form
optimal strategy, running from Nim's arithmetic rule through tic-tac-toe's tiered
tactics to the absence of any known compact rule in Connect Four and chess. The
third is the geometry of the action space, from a flat choice over a fixed cell
set, through a gravity-constrained or flanking-capture choice over a larger
board, to sliding pieces that block one another. The fourth is the planning
horizon a competent policy needs, from one move to the sixteen-move mating plan
of a two-piece endgame. A single game per level could not separate a property of
the level from an accident of the one game probing it, so at each level we build
a second, independent network for a structurally unrelated game at matched
difficulty: Wythoff's game beside Nim, Dots and Boxes beside tic-tac-toe, Othello
beside Connect Four at both board sizes, and King-and-Queen-versus-King beside
King-and-Rook-versus-King. Figure~\ref{fig:ladder} shows the resulting design.

One condition is held fixed throughout the five subsections below, and we state
it before any number is reported. Every network reported in them was
constructed by Claude Sonnet 5 under the Track A regime of
Section~\ref{sec:protocol}: full code execution, file access, parallel subagents
constructing one network each, and no limit on iteration count beyond practical
convenience. The results are therefore attributable to a model and an effort
level, and Section~\ref{sec:replication} varies both.

\subsection{Rung 0: games settled by a closed-form rule}
\label{sec:rung0}

The ladder starts where the theory of Section~\ref{sec:cost} predicts the
construction should be cheapest: games whose optimal strategy is a known
arithmetic identity. Two such games sit at this rung, and they differ in a way
the theory predicts and a single game could not reveal.

The primary game is three-pile Nim under normal play, with pile sizes bounded to
the range zero to seven so that the full state space of $8^3 = 512$
configurations is enumerable. The network is the circuit of
Figure~\ref{fig:nimcircuit}: it computes the bitwise exclusive-or of the three
pile sizes using Proposition~\ref{prop:xor}, derives Bouton's balancing target
for each pile, and scores the twenty-four candidate actions, one per
pile-and-target pair, in two tiers. The top tier fires for any action reaching
the computed balancing target; the fallback tier fires for any legal action and
is used only from positions where no balancing move exists. The replicate is
Wythoff's game over the same pile bound, which adds a move type Nim lacks,
removing the same positive amount from both piles at once, and whose optimal
strategy is the Beatty-sequence criterion of Section~\ref{sec:cost} rather than
an exclusive-or.

Both constructions are exact, and Table~\ref{tab:rung0} reports them side by
side. Of the 511 non-terminal Nim states, 448 are positions from which the player
to move can force a win, and for every one of these we verify by complete
traversal of the resulting game tree, against an opponent who may play any legal
reply at each of its turns, that the network never loses. The remaining 63 are
confirmed by an independent backward-induction solver to be genuine forced losses
against optimal opposition, from which the network is only required, and
confirmed, to produce some legal move. That same solver agrees with the
exclusive-or rule on all 512 states without exception. For Wythoff's game, seven
pairs within the bounded range are losing positions by the Beatty criterion, an
independent solver agrees with that criterion on all 64 states, and exhaustive
game-tree search confirms the network never loses from any of the 57 non-losing
states or when moving second from any of the 6 non-terminal losing ones.

\begin{table}[htbp]
\centering
\small
\begin{tabularx}{\textwidth}{@{}lrrrrX@{}}
\toprule
Game & States & Overrides & Parameters & Oracle agreement & Outcome \\
\midrule
Nim, 3 piles & 512 & 0 & 4323 & 512/512 & never loses from any of the 448 winning states \\
Wythoff, 2 piles & 64 & 57 & 3786 & 64/64 & never loses from any of the 57 non-losing states \\
\bottomrule
\end{tabularx}
\caption{Rung 0. Both constructions are exact over the complete bounded state
space, and both agree with an independently written backward-induction solver at
every state. The two differ in what exactness costs: Nim needs no override table
because its rule is an arithmetic circuit, while Wythoff's needs one entry per
state in the bounded range.}
\label{tab:rung0}
\end{table}

The qualitative finding reproduces across the two games and the quantitative one
does not, which is the point of building both. A closed-form optimal strategy
admits a complete proof at this rung under either arithmetic basis. The cost of
obtaining it differs by the whole size of the state space: Nim's circuit grows
only through the one score it must emit per candidate action, as
Proposition~\ref{prop:grundy} asserts, while Wythoff's table scales with the pile
bound, as Proposition~\ref{prop:overridecost} explains. A single Nim data point
would have reported a closed form as uniformly cheap to compile, and would have
been wrong.

\subsection{Rung 1: games decided by a two-move tactic}
\label{sec:rung1}

The second rung removes the closed form. Neither game here has an arithmetic
rule settling the move, and competence instead rests on a short list of tactical
conditions, the deepest of which spans two moves. The question is whether a
tiered scorer plus a table of exceptions still closes the gap to optimal play
once no single identity does the work.

The primary game is tic-tac-toe, the one network reused rather than built for
this study, and the same one used as the worked example in every brief. Its
construction is a tiered heuristic of win, block, fork, block-fork, centre,
opposite corner, corner and side, topped with 80 exact-match override entries for
the reachable positions where that single-pass reasoning is insufficient. The
replicate is Dots and Boxes on a two-by-two grid of boxes, with twelve possible
edges and four boxes, which replaces tic-tac-toe's cell-based action space and
strict alternation with an edge-based action space and a turn-extension rule, so
that completing a box grants an immediate extra move. This is the first
action-space change in the ladder that is not a restriction on which cells are
legal.

Both are exact at every reachable state, and Table~\ref{tab:rung1} reports them.
The substantive finding is in the override column. Dots and Boxes needs 556
entries, 13.6 per cent of its non-terminal states, against tic-tac-toe's 80
entries, 1.8 per cent, and the reason is the classical double-cross phenomenon:
declining an available box completion, to force a more favourable chain split, is
sometimes strictly better than taking the box. The rule "complete a box when
possible" is therefore not the unconditionally correct top tier it would be in a
game where a terminal-in-one-move tactic is always right. Accommodating this
required moving the override tier above the box-completion tier in the scoring
hierarchy.

\begin{table}[htbp]
\centering
\small
\begin{tabularx}{\textwidth}{@{}lrrrX@{}}
\toprule
Game & Non-terminal states & Overrides & Parameters & Outcome \\
\midrule
Tic-tac-toe, $3{\times}3$ & 4520 & 80 (1.8\%) & 15\,781 & never loses moving first or second \\
Dots and Boxes, $2{\times}2$ & 4095 & 556 (13.6\%) & 16\,924 & matches the exact game value at every state \\
\bottomrule
\end{tabularx}
\caption{Rung 1. Both constructions are exact under exhaustive game-tree search
against an adversarial opponent. The seven-fold difference in override share
measures how much of each game's optimal play the hand-derived tiers fail to
capture, and it is the double-cross tactic in Dots and Boxes that accounts for
the gap.}
\label{tab:rung1}
\end{table}

The recipe survives an action-space change with no counterpart in tic-tac-toe,
and one belief that tic-tac-toe alone would have supported does not. The claim
that the top heuristic tier is always safe to apply unconditionally is a property
of tic-tac-toe's particular tactics and not of this difficulty level. This is the
first of several places where the replicate corrects a generalization the primary
game invited.

\subsection{Rung 2a: enumerable boards with a new action geometry}
\label{sec:rung2a}

The third rung changes the geometry of the action space while keeping the state
space enumerable, so that exact proof remains available and the effect of the
geometry can be isolated. Board sizes were chosen empirically as the largest
whose full game tree remains exhaustively checkable.

The primary game is Connect Four on a four-column, five-row board, winning on
three in a row, which introduces gravity: a move names a column and the piece
falls. The board has 195\,472 reachable states, of which 104\,399 are
non-terminal. Because this board is a forced win for whoever moves first rather
than a theoretical draw, the correctness claim proved here is stronger than
tic-tac-toe's. We prove not merely that the network never loses, which would be
false whenever it moves second, but that from every one of the 104\,399
non-terminal states, against an opponent who may choose any legal reply, the
network achieves exactly the game-theoretic value an independent minimax oracle
assigns to that state, whether that value is a win, a draw or a forced loss. The
replicate is Othello on a four-by-four board, which replaces gravity with
flanking capture: a legal move lands on any empty cell that outflanks at least
one opposing line in any of eight directions. It has 53\,651 non-terminal
reachable states, of which 45\,387 have at least one legal move, the remaining
8264 being positions where the correct action is to pass, a rule with no
counterpart in Connect Four.

Both constructions are exact at every reachable state, including every forced
pass, and Table~\ref{tab:rung2a} reports them. Two measured facts are worth
separating from the headline. The first is that this Othello board's initial
position is a loss for whoever moves first under optimal play on both sides,
which the network reproduces exactly and which we did not assume. The second is
the override share: Othello's table absorbs 2537 of the 45\,387 states where a
move exists, 5.6 per cent, against Connect Four's 751 of 104\,399, 0.72 per cent,
a ratio of nearly eight. Othello has no terminal-in-one-move tactic analogous to
completing a line, so one-ply greedy reasoning sees proportionally less of what
the exact table ultimately supplies.

\begin{table}[htbp]
\centering
\small
\begin{tabularx}{\textwidth}{@{}lrrX@{}}
\toprule
Game & States & Overrides & Outcome \\
\midrule
Connect Four, $4{\times}5$ & 104\,399 & 751 (0.72\%) & exact game value from every state; first-player win reproduced \\
Othello, $4{\times}4$ & 53\,651 & 2537 (5.6\%) & exact game value from every state, including all 8264 forced passes \\
\bottomrule
\end{tabularx}
\caption{Rung 2a. The state counts are non-terminal reachable positions, and the
override shares are taken over the positions at which the network must choose a
move, 104\,399 for Connect Four and the 45\,387 active positions for Othello.
Both boards are solved exactly, at a scale respectively twenty-three and twelve
times tic-tac-toe's. The override share differs by nearly a factor of eight
between the two mechanics, which measures how much of optimal play a one-ply
tactical vocabulary captures in each. The two networks hold 62\,363 and
139\,361 parameters respectively.}
\label{tab:rung2a}
\end{table}

The tiered-heuristic-plus-override pattern therefore survives two unrelated
changes of action geometry intact, at a cost that varies with how much of the
game's tactics are visible one move ahead. Connect Four's larger state space and
smaller table together show that the table's size tracks the tactical texture of
the game and not its size, which is the empirical counterpart of
Proposition~\ref{prop:overridecost}.

\subsection{Rung 2b: standard boards, where proof is replaced by measurement}
\label{sec:rung2b}

The fourth rung holds the rules and the geometry fixed and enlarges the board
until the state space is no longer enumerable. No override table can be
constructed, no exact proof is available, and the network is the tiered heuristic
alone. This is the rung at which the study can run a controlled comparison it
cannot run anywhere else, because the opponent can be given the same kind of
evaluation as the network and differ from it in exactly one respect.

Connect Four on the standard seven-by-six board has a reachable state space of
order $10^{13}$. The resulting network implements the first three of rung 2a's
tiers plus a static centre-column bias, with fork detection again omitted.
Othello on the standard eight-by-eight board carries its five tiers unchanged.
Both are evaluated across ten random seeds and 400 games per condition and side,
against two opponents: one choosing uniformly at random, and one running an
independently written alpha-beta search with a hand-rolled evaluation sharing no
code with the network. Table~\ref{tab:rung2b} reports the outcome.

\begin{table}[htbp]
\centering
\small
\begin{tabularx}{\textwidth}{@{}lrrrX@{}}
\toprule
Game & Parameters & vs.\ random & vs.\ search & Search opponent \\
\midrule
Connect Four, $7{\times}6$ & 251\,477 & 99.5--99.8\% & 0/800 & depth-4 alpha-beta \\
Othello, $8{\times}8$ & 1\,527\,080 & 86.3--87.5\% & 0/800 & depth-3 alpha-beta \\
\bottomrule
\end{tabularx}
\caption{Rung 2b. Ten seeds and 400 games per condition and side. The two ranges
against random play are the first-moving and second-moving rates. Against search,
both networks lose every game in both seats, with no exception across 1600 games
in total.}
\label{tab:rung2b}
\end{table}

The pattern is the same under gravity and under flanking capture, and it is
sharp. Against an opponent with no planning at all, the Connect Four network wins
99.5 per cent of games moving second and 99.8 per cent moving first, and the
Othello network wins 86.3 per cent moving first and 87.5 per cent moving second.
The visible gap between the two reflects Othello's lack of a terminal-in-one-move
tactic, which is the same property that produced its larger override table at
rung 2a. Against an opponent with even a small amount of planning, both lose
every single game. A depth-four search is a weak Connect Four player by the
standards of the solved game, and it still wins 800 games out of 800.

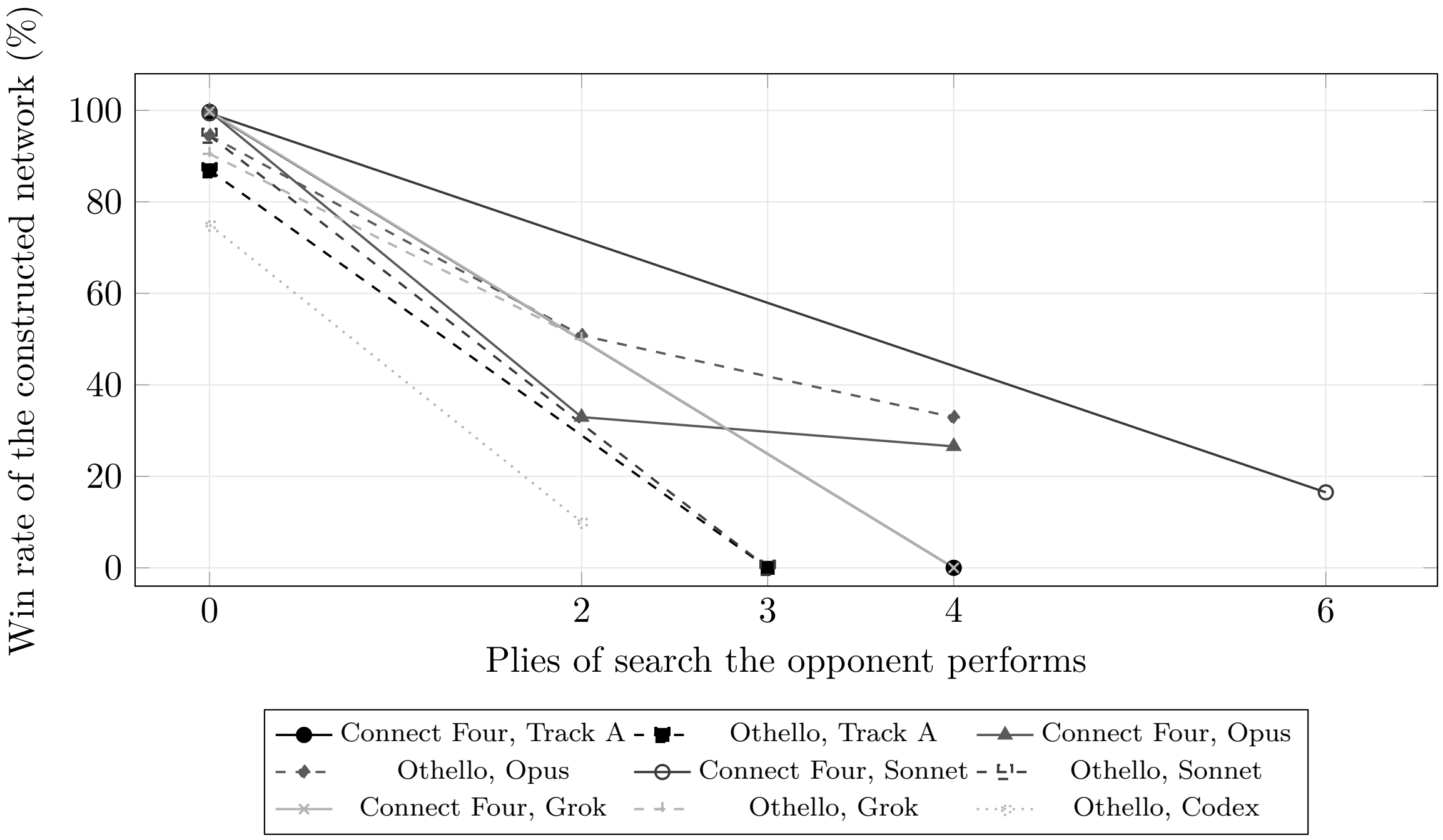
\begin{figure}[htbp]
\centering
\begin{tikzpicture}
\begin{axis}[
  width=0.93\linewidth,
  height=7.0cm,
  xlabel={Plies of search the opponent performs},
  ylabel={Win rate of the constructed network (\%)},
  xmin=-0.4, xmax=6.6,
  ymin=-4, ymax=108,
  xtick={0,2,3,4,6},
  ytick={0,20,40,60,80,100},
  grid=both,
  grid style={gray!18},
  legend style={at={(0.5,-0.24)}, anchor=north, legend columns=3, font=\scriptsize},
  every axis plot/.append style={mark size=2.1pt, line width=0.7pt},
]
\addplot[black, mark=*] coordinates {(0,99.65) (4,0)};
\addlegendentry{Connect Four, Track A}
\addplot[black, dashed, mark=square*] coordinates {(0,86.9) (3,0)};
\addlegendentry{Othello, Track A}
\addplot[gray!70!black, mark=triangle*] coordinates {(0,99.80) (2,32.95) (4,26.55)};
\addlegendentry{Connect Four, Opus}
\addplot[gray!70!black, dashed, mark=diamond*] coordinates {(0,94.50) (2,50.80) (4,32.90)};
\addlegendentry{Othello, Opus}
\addplot[gray!45!black, mark=o] coordinates {(0,99.35) (6,16.5)};
\addlegendentry{Connect Four, Sonnet}
\addplot[gray!45!black, dashed, mark=square] coordinates {(0,94.45) (3,0)};
\addlegendentry{Othello, Sonnet}
\addplot[gray!60, mark=x] coordinates {(0,99.75) (4,0)};
\addlegendentry{Connect Four, Grok}
\addplot[gray!60, dashed, mark=+] coordinates {(0,90.5) (2,50.0)};
\addlegendentry{Othello, Grok}
\addplot[gray!60, dotted, mark=asterisk] coordinates {(0,75.0) (2,10.0)};
\addlegendentry{Othello, Codex}
\end{axis}
\end{tikzpicture}
\caption{Every measurement in this paper of a constructed network against an
opponent performing a stated number of plies of search, assembled on one axis.
Depth zero is a uniformly random opponent. Each line joins measurements of one
construction, and the lines are not a single curve: each attempt's search
opponent carries that attempt's own hand-set evaluation, so the vertical spread
at a given depth mixes opponent strength with construction quality, and no
attempt varied depth with everything else held fixed. What survives that caveat
is the direction. Every construction is at or near dominance against an opponent
with no planning, every one of them falls as planning is added, and the four that
were tested against the deepest opponent their builder chose reach or approach
zero. Measuring the curve properly, under one construction and one evaluation
with only depth varying, is the clearest piece of future work this study leaves.}
\label{fig:depthcurve}
\end{figure}

What makes this rung a controlled result rather than another data point is that
the only quantity separating the two opponents is the presence of recursive
search. The margin against the non-searching opponent depends on the game's
tactical texture; the margin against the searching one does not depend on
anything we varied. Figure~\ref{fig:depthcurve} assembles every measurement of
this kind in the paper, including those from the independent attempts of
Section~\ref{sec:replication}, against the opponent's search depth.

\subsection{Rung 3: chess endgames, where local correctness stops composing}
\label{sec:rung3}

The last rung keeps three pieces on the board and changes two axes at once,
because no game we could find changes only one. The action space becomes sliding
and mutually blocking, so the legality of a rook's destination depends on where
the rook currently stands, an input-dependent quantity rather than a fixed board
cell. And competent play requires a plan of double-digit length rather than a
tactic spanning one or two moves. The input is 192 binary dimensions, one-hot
over 64 squares for each of three pieces.

The primary game is King-and-Rook against a lone king. The construction builds
the rook's and king's legal-move indicators as an input-conditioned conjunction
over every geometrically aligned origin square, using the exact-match and
clipping gadgets of Section~\ref{sec:gates} together with fixed board-geometry
constants, namely which squares are mutually visible along a shared rank or file
and which are Chebyshev-adjacent. It then scores candidate moves in tiers
favouring immediate checkmate, avoiding stalemate, approaching with the rook
along a shrinking confinement box when that approach is safe from capture, and
otherwise approaching with the attacking king. The replicate is
King-and-Queen against a lone king, which uses a piece with the rook's four
sliding directions plus the bishop's four diagonals and, unlike the rook
construction, a dedicated circuit computing the opponent's exact post-move
mobility for every candidate move, which converts the mate-in-one and
stalemate-avoidance detectors from incidental to exact.

The oracles are validated before anything is checked against them. The
King-and-Rook oracle computes 175\,168 legal positions with the network to move,
223\,944 with the lone king to move, 22\,244 positions that are already a
stalemate or allow the rook to be captured for free, and a maximum of sixteen
moves to force mate from any winning position with best play on both sides. All
three position counts agree exactly with the independent public computation of
\citet{leslie-hurd-krk}, and the maximum of sixteen agrees exactly with the value
first computed in Str\"ohlein's thesis as reported by
\citet{haworth2020strohlein}. The King-and-Queen oracle reproduces that account's
maximum of ten moves exactly, alongside 144\,508 legal positions with the network
to move.

Table~\ref{tab:rung3} reports both networks against those oracles, and the
columns are arranged so that the collapse is visible along a row. Each network is
checked at every legal position with no sampling, and then simulated over 3000
complete games against an opponent that selects, at each of its turns, the reply
worst for the network, under a cap of forty moves for the rook endgame and a cap
three times the true worst case for the queen endgame.

\begin{table}[htbp]
\centering
\small
\begin{tabularx}{\textwidth}{@{}lrrrrX@{}}
\toprule
Endgame & Legal & Mate in one & Strict progress & Full game & Dominant failure mode \\
\midrule
K+R vs.\ K & 100\% & 84.9\% (1284/1512) & 6.99\% & 0.57\% & rook lost for nothing, in 58.5\% of games \\
K+Q vs.\ K & 100\% & 100\% (2448/2448) & 39.06\% & 9.13\% & move cap exhausted, in 90.87\% of games \\
\bottomrule
\end{tabularx}
\caption{Rung 3, measured over the 175\,168 and 144\,508 legal positions with
the network to move. Every column except the last two is exhaustive over that
complete set, with no sampling. Mate in one is the share of positions at oracle
distance one from mate at which the network delivers it; strict progress is the
share of all positions at which the chosen move strictly decreases the oracle's
distance to mate; full game is the share of 3000 simulated games ending in
checkmate against an opponent playing the worst available reply at every turn.
The queen endgame additionally records zero avoidable stalemates and zero
unforced queen captures across all 144\,508 positions and all 3000 games.}
\label{tab:rung3}
\end{table}

The single most informative fact at this rung is the distance between the third
column and the sixth. The rook network's board-geometry reasoning is exact, it
never proposes an illegal move in 175\,168 out of 175\,168 positions, and its
one-ply tactical judgment is sound in 84.9 per cent of the positions where an
immediate mate exists. Yet its chosen move strictly decreases the distance to
mate in only 6.99 per cent of positions, and a per-move progress rate of one in
fourteen, compounded over a plan needing on the order of sixteen consecutive
correct decisions, produces near-total failure the moment the opponent is
adversarial rather than passive. In 3000 games it reaches checkmate 0.57 per cent
of the time, loses the rook for no compensation in 58.5 per cent, and exceeds the
forty-move cap without resolution in a further 40.9 per cent.

The replicate separates two questions the rook endgame leaves conflated: whether
one-ply safety guarantees compose into full-game safety, and whether one-ply
progress guarantees compose into a finished plan. On the first the answer is now
almost perfectly yes. The queen network allows no avoidable stalemate and no
unforced capture at any of its 144\,508 positions, and loses the queen in none of
3000 games, against the rook endgame's 58.57 per cent combined failure on the
same two outcomes. On the second it remains largely no: 90.87 per cent of games
exhaust a move cap three times the true worst case without reaching mate,
comparable in kind to the rook endgame's 99.43 per cent non-mate rate. The
severity of the rung is therefore not one fixed quantity. The safety axis scales
sharply with whether the one-ply checks involved are exact or merely heuristic,
and the planning axis improves only modestly under a shorter required plan,
rising from 0.57 to 9.13 per cent. Neither distinction is visible from either
network studied alone.

\subsection{What the ladder cost to build}
\label{sec:cost-measured}

Before varying the constructor it is worth collecting the ten networks in one
place, since the construction cost is itself a measurement and it behaves in a
way that the per-rung tables above do not show. Table~\ref{tab:summary} gives
each network's verification standard and headline outcome, and
Table~\ref{tab:cost} gives its size, its override table and the wall-clock time
of its full verification.

\begin{table}[htbp]
\centering
\small
\begin{tabularx}{\textwidth}{@{}llcX@{}}
\toprule
Rung & Game & Verification & Headline outcome \\
\midrule
0 & Nim, 3 piles & exhaustive, 512 states & never loses from any winning position \\
0, replicate & Wythoff's game, 2 piles & exhaustive, 64 states & never loses from any winning position \\
1 & Tic-tac-toe & exhaustive, 4520 states & never loses moving first or second \\
1, replicate & Dots and Boxes, $2{\times}2$ & exhaustive, 4095 states & matches exact game value from every state \\
2a & Connect Four, $4\times5$, connect-3 & exhaustive, 104\,399 states & matches exact game value from every state \\
2a, replicate & Othello, $4\times4$ & exhaustive, 53\,651 states & matches exact game value from every state \\
2b & Connect Four, $7\times6$ & statistical, 1600 games & 99.5--99.8\% vs.\ random; 0/800 vs.\ depth-4 search \\
2b, replicate & Othello, $8\times8$ & statistical, 1600 games & 86.3--87.5\% vs.\ random; 0/800 vs.\ depth-3 search \\
3 & Chess endgame, K+R vs.\ K & exact oracle, 175\,168 states & 100\% legal; 84.9\% tactic; 0.57\% full game \\
3, replicate & Chess endgame, K+Q vs.\ K & exact oracle, 144\,508 states & 100\% legal; 100\% tactic; 9.13\% full game \\
\bottomrule
\end{tabularx}
\caption{Verification standard and headline outcome for all ten networks.
Replicate rows share every difficulty axis with the primary game above them and
use a structurally unrelated mechanic.}
\label{tab:summary}
\end{table}

\begin{table}[htbp]
\centering
\scriptsize
\begin{tabularx}{\textwidth}{@{}Xrrrrl@{}}
\toprule
Network & Parameters & Input/output dims & Override table & State space & Wall-clock \\
\midrule
Nim & 4323 & 9 / 24 & 0 (arithmetic) & 512 & $\sim$1\,s \\
Wythoff & 3786 & 6 / 22 & 57 (full table) & 64 & $\sim$1\,s \\
Tic-tac-toe & 15\,781 & 18 / 9 & 80 & 4520 & $\sim$2\,s \\
Dots and Boxes & 16\,924 & 12 / 12 & 556 & 4095 & $\sim$5\,s \\
Connect Four, small & 62\,363 & 40 / 4 & 751 & 104\,399 & $\sim$100\,s \\
Othello, small & 139\,361 & 32 / 16 & 2537 & 53\,651 & $\sim$60\,s \\
Connect Four, full & 251\,477 & 84 / 7 & n/a (intractable) & $\sim10^{13}$ & $\sim$190\,s \\
Othello, full & 1\,527\,080 & 128 / 64 & n/a (intractable) & intractable & $\sim$15\,min \\
KRK endgame & 2\,352\,516 & 192 / 128 & n/a & 175\,168 & $\sim$85\,s \\
KQK endgame & 38\,898\,084 & 192 / 128 & n/a & 144\,508 & $\sim$6\,min \\
\bottomrule
\end{tabularx}
\caption{Construction-cost measurements for every network: total hand-set
parameter count, input and output tensor dimensions, size of the exact-match
override table where one is used, the independent oracle's state-space size, and
wall-clock time for this paper's full verification protocol on the development
machine, which has sixteen cores and no graphics accelerator. Statistical
timings include the full seeded evaluation script. Every parameter count in this
table was recomputed directly from the module's own parameter list for this
version of the manuscript, and every network holds zero trainable parameters.}
\label{tab:cost}
\end{table}

Parameter count spans four orders of magnitude across the ladder, and it is not
board size that drives it. The queen endgame is the largest network in the study
at 38\,898\,084 parameters, sixteen times the rook endgame's, on an identical
input encoding and an identical board; the difference is that its tactical tiers
are exact circuits computing the opponent's post-move mobility, where the rook
network's tiers are unconditional heuristics whose correctness at a given
position is incidental. Exactness in a tier, not size in a board, is what costs
parameters. The same comparison at rung 0 points the other way: Wythoff's network
is the smallest in the study and is exact, because the state space it tabulates
is tiny. Cost in this paradigm is governed by the product of how exact a tier has
to be and how many positions it cannot cover.

\subsection{Varying the constructor}
\label{sec:replication}

Every number above was produced by one model under one regime, so it reports a
capability of that model as much as a property of the technique. This subsection
separates the two. We released the task specification, stripped of every
gate-construction hint specific to our own solutions, as a self-contained brief,
and had other models attempt the same networks independently in isolated
directories under the same full-tool-access regime. Six such attempts are
reported here: two administered by a collaborator using models from other
providers, and three spanning one provider's own model sizes, plus one
higher-effort repeat. We take the external attempts first, then the
within-family comparison, and we state for each what we re-verified ourselves
rather than accepted on report.

The two external attempts were obtained from two different coding agents working from the identical brief in two separate isolated directories, with no access to each other's output. The first, reported to us by the collaborator administering it as "6 Luna High" (a label we reproduce as given, since we did not ourselves verify the exact underlying model identity), is referred to below as the Codex replication. The second, reported to us as Grok 4.7 High and referred to below as the Grok replication, worked from the same brief independently and finished after this section was first drafted; we verified it before extending the section rather than taking its own account on faith. Both replicating agents worked in a directory isolated from this paper's own source code and were instructed not to import, read, or otherwise reuse any of our oracles, reference opponents, or network implementations; for the Grok replication we additionally confirmed this by grepping its entire codebase for references to this paper's own module paths and finding none, and by independently re-running its full test suite and its two exhaustive 223\,944-position chess-endgame checks ourselves rather than trusting its own reported pass/fail status. Each agent's own account of its construction and every limitation it encountered is in its own \texttt{SUMMARY.md} (and, for Codex, \texttt{METHOD.md}), and every number we report below was cross-checked by us against each replication's raw per-task \texttt{RESULTS.md} files and result JSON, not taken from either agent's summary prose alone. The brief's eleventh network, general chess, is outside this paper's own four-rung scope; most attempts built and statistically evaluated it regardless, and we do not discuss those results further here.

The replication's exact-proof tier reproduces our own most directly on the four rungs where the state space is small enough to enumerate under either model's choice of bound: Nim (512 of 512 bounded configurations, matching ours exactly), Wythoff's game (441 of 441 states), the reduced Connect Four board (104\,413 reachable positions, 104\,399 of them non-terminal, matching our own 104\,399 non-terminal count), and the reduced Othello board (61\,619 reachable relative states, against our own 53\,651). Two details are worth isolating from the headline agreement. First, the replicating agent bounded Wythoff's piles to the full zero-to-twenty range the original task specification states, where our own construction, after an internal inconsistency documented in its own \texttt{RESULTS.md}, ended up bounding piles to zero-to-seven instead (Table~\ref{tab:cost}); the replication is therefore the one attempt, of the two, that followed the written specification literally on this task. Second, the two attempts' Othello state counts and measured initial-position values differ (53\,651 states and a value of $-1$ for us, 61\,619 states and a value of $-8$ for the replication); the sign of both values agrees, a loss for whoever moves first, but the magnitude disagreement traces to the replication's oracle reporting the exact final disc differential under optimal play rather than a bare win-loss-draw sign, a finer-grained and compatible notion of game value.

On Dots and Boxes, the replication surfaces a genuine limitation of the task specification itself rather than a construction failure: it reports that the specified 12-bit edge-mask input does not by itself disambiguate every reachable position, since an exhaustive history audit on its side found 875 of 4096 edge masks reachable from more than one combination of score and side-to-move. Declining to claim exact play it could not verify under that input contract, it instead implements and reports a non-game-theoretic deterministic baseline (always draw the lowest-index undrawn edge), which we evaluate on its own terms rather than treat as a failed attempt at the stronger guarantee: 11 wins, 2 draws, and 7 losses in 20 games against a uniformly random opponent (55\%, Wilson 95\% CI [0.342, 0.742]), and 1 win, 1 draw, and 18 losses in 20 games against an independently written perfect-information minimax opponent (5\%, CI [0.009, 0.236]). We did not independently verify whether the ambiguity it identifies actually changes the optimal next move at an affected mask, since the value of continuing play from a given edge configuration is, by the rules of the game, determined by the remaining edges and whose turn it is rather than by the score already banked; we report the finding and our own construction's exact result side by side in Table~\ref{tab:replication} rather than adjudicate between them here.

\begin{table}[htbp]
\centering
\tiny
\begin{tabularx}{\textwidth}{@{}lXXX@{}}
\toprule
Task & Claude Sonnet 5 & Codex (``6 Luna High'') & Grok (Grok 4.7 High) \\
\midrule
Nim & exhaustive, 512 states, exact & exhaustive, 512 states, exact & exhaustive, 512 states, exact \\
Wythoff & exhaustive, 64 states (piles 0--7) & exhaustive, 441 states (piles 0--20, as specified) & exhaustive, 441 states (piles 0--20, as specified) \\
Dots and Boxes & exhaustive, 4095 states, exact & representation ambiguity identified (875/4096 masks); non-optimal baseline, 55\% vs.\ random, 5\% vs.\ minimax (20 games each) & exhaustive, 4095 states, exact; 0 score/edge conflicts found in 5101 reachable mask-score pairs \\
Connect Four, $4{\times}5$ & exhaustive, 104\,399 states, exact & exhaustive, 104\,413 states (104\,399 non-terminal), exact & exhaustive, 104\,399 non-terminal states, exact \\
Othello, $4{\times}4$ & exhaustive, 53\,651 states, exact, initial value $-1$ & exhaustive, 61\,619 states, exact, initial value $-8$ (disc differential) & exhaustive, 58\,533 states, exact, initial value $-1$ \\
Connect Four, $7{\times}6$ & 99.5--99.8\% vs.\ random, 0/800 vs.\ depth-4 (10 seeds $\times$ 400 games) & 100\% vs.\ random, 0\% vs.\ depth-3 (10 seeds $\times$ 2 games) & 99.75\% vs.\ random (400 games), 0\% vs.\ depth-4 (80 games) \\
Othello, $8{\times}8$ & 86.3--87.5\% vs.\ random, 0/800 vs.\ depth-3 & 75\% vs.\ random, 10\% vs.\ depth-2 (20 games each) & 90.5\% vs.\ random (200 games), 50\% vs.\ depth-2 (40 games) \\
KRK endgame & exact oracle, 175\,168 states; 0.57\% full game vs.\ worst-case adversary & not a literal weight-only circuit (self-reported); 5\% vs.\ random king, 0\% vs.\ 1-ply search (20 games each) & literal ReLU circuit, exact oracle, 175\,168 legally reachable of 223\,944 enumerated states (exact match to our own count); 41.5\% vs.\ random king, 1.0\% vs.\ optimal defense (200 games each) \\
KQK endgame & exact oracle, 144\,508 states; 9.13\% full game vs.\ worst-case adversary & not a literal weight-only circuit (self-reported); 0\% vs.\ random king, 0\% vs.\ 1-ply search (20 games each) & literal ReLU circuit, exact oracle, 144\,508 legally reachable of 223\,944 enumerated states (exact match to our own count); 19.5\% vs.\ random king, 10.0\% vs.\ optimal defense (200 games each) \\
\bottomrule
\end{tabularx}
\caption{This paper's Track A result beside both completed independent replications, task by task. Sample sizes differ across all three attempts (noted inline), which affects the width of each replication's confidence intervals relative to our own; each replication's own self-reported scope limitations are carried over verbatim rather than smoothed into a single comparable number. Grok's KRK/KQK oracle enumerates 223\,944 piece placements before filtering out those in which the black king is already in check, a configuration illegal with white to move; filtering that condition out ourselves, directly from Grok's own code, leaves exactly 175\,168 and 144\,508 placements respectively, matching our own two endgames' independently computed legal-position counts exactly. Othello's three attempts use three different state-counting conventions (53\,651, 58\,533, and 61\,619 reachable positions respectively) and, for Codex, a different game-value convention (exact disc differential rather than win-loss sign); all three agree on the sign of the initial position's value. The Othello, $8\times8$ row's Grok result against a two-ply search opponent, 50\%, is the one result across either replication that does not reproduce this paper's otherwise uniform collapse against search, discussed in Section~\ref{sec:replication}.}
\label{tab:replication}
\end{table}

On the two statistically evaluated rungs with no exact-proof claim at either scale, Connect Four and Othello, the replication used ten seeds but only two games per seed and condition (20 games per condition, against our own 400 for both), which widens every one of its confidence intervals substantially; within that smaller sample, it reports 100\% against a random Connect Four opponent and 0\% against an independent depth-3 search, and 75\% against a random Othello opponent and 10\% against an independent depth-2 search. Every one of these four numbers reproduces our own study's qualitative pattern, a heuristic that beats a non-planning opponent and loses to any opponent with real search, under a third independently constructed set of weights.

On the two chess endgames, the replicating agent's own account is the most useful summary, and we reproduce its substance rather than soften it: it states explicitly that its King-and-Rook and King-and-Queen modules are "not a literal all-weights-in-Linear-layers construction" but a one-ply geometric heuristic scored in ordinary Python control flow, and that this should be treated as a method limitation rather than counted as a successful pure weight-space construction under this paper's Definition~\ref{def:wsp}. Measured on its own terms nonetheless, on a 200-position sampled legality and stalemate-avoidance audit (not the exhaustive full-state-graph check Section~\ref{sec:experiments} performs) it found no illegal move and no avoidable stalemate; in 20 simulated games per condition it checkmated a uniformly random lone king once for King-and-Rook (5\%) and never for King-and-Queen (0\%), and never checkmated an independent one-ply search opponent for either endgame (0\% in both cases). These opponents are weaker than the worst-case adversarial opponent our own 0.57\% and 9.13\% full-game figures are measured against, which makes the comparison asymmetric rather than directly quantitative, but the direction is unambiguous: a construction that does not satisfy this paper's weight-space-programming definition fails to force mate reliably even against opponents easier than the ones our own, definition-compliant networks were tested against.

The Grok replication's exact-proof tier covers the same four enumerable rungs as the Codex replication, plus Dots and Boxes, which Codex had declined to claim exactly; on all five, Grok's numbers either match our own or match Codex's, and on one task the three-way comparison itself is informative. Nim agrees with both other attempts exactly, 512 of 512 bounded configurations. Wythoff's game again uses the full zero-to-twenty bound the task specification states, 441 of 441 states, matching Codex rather than our own zero-to-seven bound and reinforcing, now from two independent attempts rather than one, that our own bound was the non-literal reading of the specification. The reduced Connect Four board matches our own non-terminal count exactly, 104\,399 positions, with the same first-player-win value, a three-way exact agreement among all three attempts on this specific number. The reduced Othello board again returns a fourth distinct total state count, 58\,533 reachable positions of which 45\,387 have a legal move, between our own 53\,651 and Codex's 61\,619, but its value convention matches our own rather than Codex's, $-1$ as a bare win-loss sign for the second player rather than Codex's $-8$ disc differential; three different state-counting conventions and a shared outcome sign is the pattern this task now shows across three independent constructions. On Dots and Boxes, Grok reports an exact solve and explicitly reports having tested for the kind of ambiguity Codex raised: it enumerates all 5101 reachable pairs of edge-mask and mover score and finds the set of optimal edges at a given mask never empty across scores, zero conflicts, agreeing with our own exact result rather than with Codex's. We did not adjudicate this disagreement ourselves in the previous paragraph and do not do so now; we note only that two of the three independent constructions, including the one that explicitly checked for the ambiguity Codex reports, found none.

On the statistically evaluated rungs, Grok used substantially larger samples than Codex, closer in scale to our own: 400 games against random play and 80 against an independent depth-4 search for standard Connect Four (matching our own condition counts almost exactly), and 200 against random play for standard Othello. Connect Four reproduces our pattern cleanly, 99.75\% against random (Wilson 95\% CI [0.986, 1.000]) and exactly 0\% in 80 games against depth-4 search (CI [0, 0.046]). Standard Othello's random-opponent result also reproduces ours closely, 90.5\% (CI [0.856, 0.938]) against our own 86.3--87.5\%. Its search result does not: against a 40-game sample at search depth two, one ply shallower than the depth-three opponent our own Othello network was tested against, Grok's network scores 50\% (20 wins, 0 draws, 20 losses, CI [0.352, 0.648]), an even fight rather than the near-total collapse every other heuristic-versus-search comparison in both this paper and both replications shows. This is the one genuine exception to an otherwise uniform pattern across ten primary networks and two independent replications of both of them, and we report it rather than smooth over it: it suggests the collapse this paper documents sets in at a specific, small, nonzero search depth rather than at literally any search depth greater than zero, and that for this particular game and this particular pair of hand-set evaluation functions, that threshold sits somewhere between two and three plies rather than at one. We revisit this exception below: a later, higher-effort rebuild of the same model on the same task does not reproduce it, which weakens, without eliminating, the case for treating it as a stable fact about this game rather than about this one construction.

On the two chess endgames the Grok replication is the more direct test of this
paper's central claim, because its King-and-Rook and King-and-Queen modules are,
like our own, literal \texttt{nn.Linear}-and-\texttt{ReLU} circuits rather than
procedural heuristics. We confirmed that ourselves, by inspecting both modules'
source and by recomputing each network's parameter count directly from
\texttt{net.parameters()}. Both endgames use an exact legality, mate-in-one,
stalemate and hanging-piece circuit built from the gate primitives of
Section~\ref{sec:gates}, plus an additive geometric confinement bonus the agent
itself describes as a heuristic and not a shortest-mate policy.

Its oracle counts positions differently from ours, and reconciling the two
conventions produces the sharpest agreement in the paper. Grok's oracle
enumerates 223\,944 piece placements for each endgame, namely every triple of
squares with the two kings not Chebyshev-adjacent and no two pieces sharing a
square. It solves and checks all of them, without filtering out the placements in
which the black king already stands in check, a configuration that cannot legally
arise with white to move. We applied that filter ourselves, directly to Grok's
own oracle code. Of the 223\,944 placements, exactly 175\,168 have the black
king not in check for King-and-Rook and exactly 144\,508 for King-and-Queen,
matching our own legal-position counts to the last digit under an otherwise
independently written retrograde solver. We then checked both networks
exhaustively over all 223\,944 placements. Every action scoring above each
network's legality threshold is exactly the legal move set, and the argmax is a
mate in one whenever one exists. That holds on the legally reachable placements
and on the 48\,776 and 79\,436 additional placements that are well-defined
inputs without being reachable positions. The solver behind the check agrees with
ours on both maxima, sixteen moves and ten.

The game results reproduce our own pattern and one of them lands remarkably close
to it. Against a uniformly random lone king, King-and-Rook mates 41.5 per cent of
the time (83 of 200, CI [0.349, 0.484]) and King-and-Queen 19.5 per cent (39 of
200, CI [0.146, 0.255]). Against each endgame's own exact retrograde policy
playing optimal defence, the rates fall to 1.0 per cent (2 of 200, CI [0.003,
0.036]) and 10.0 per cent (20 of 200, CI [0.066, 0.149]). That last figure, from
an independently built literal circuit, lands within one percentage point of our
own King-and-Queen network's 9.13 per cent against a worst-case adversary.
Neither construction was tuned to produce the agreement, and the two sets of
weights were written by two different models from the same specification. Neither
replication network stalemates the lone king or hangs its own piece in any of the
400 combined games, matching our networks' one-ply safety record exactly, and
neither reaches a majority mate rate against optimal defence, matching our
networks' failure to compose that safety into a finished plan.

Both replicating agents were later asked, by the same collaborator and under the same two isolated directories, to repeat the identical brief at a higher reported effort setting, which lets us ask whether raising effort within one model changes where the collapse this paper documents sets in, holding the model itself fixed. We treat the two resulting attempts very differently, because one of them disclosed, in its own summary, that it was not an independent attempt at all: the higher-effort Codex run states plainly that it "reruns the first run's implementation recipe\ldots{} It is a reproducibility run, not an independent rewrite of the source strategy," and its numbers are correspondingly near-identical to its own earlier run's, task by task. Its working directory also had standing read access to this paper's own source tree at the operating-system level, which the first Codex attempt did not, so even absent that disclosure we would not have treated this run as independent. We exclude it from every comparison in this paper for that reason, stated here rather than left implicit.

The higher-effort Grok run shows no such sign of reuse: its code is independently isolated from every other attempt in this study by the same grep-based check applied throughout this section, and, more tellingly, its own numbers and even its own choice of network size differ substantially from its own earlier run's, exactly what a genuine second construction attempt should produce and a mere rerun should not. Its King-and-Rook and King-and-Queen networks use 2\,367\,140 and 2\,335\,192 parameters, each recomputed by us from \texttt{net.parameters()}, against 62\,031\,859 and 62\,014\,867 at the earlier effort setting. That is a twenty-six-fold reduction, and the result remains a literal \texttt{nn.Linear}-and-\texttt{ReLU} circuit throughout. We confirmed it exhaustively over all 223\,944 enumerated placements per endgame, as before, and the 175\,168 and 144\,508 legally reachable among them again match our own counts exactly. Against a uniformly random lone king it now mates 78.5\% of the time for King-and-Rook and 19.5\% for King-and-Queen; against each endgame's own optimal retrograde defender, 2.5\% and 14.0\% respectively, both modest improvements over the earlier run's 1.0\% and 10.0\%, consistent with more effort helping somewhat without moving either endgame off the same side of the architectural ceiling. The Othello exception does not survive this rebuild: against the same two-ply search opponent that produced the earlier run's 50\%, the higher-effort network loses every one of 40 games, 0\%, restoring the uniform collapse pattern. We read this as evidence that the earlier 50\% result reflects variance in one particular construction's one-ply heuristic rather than a stable fact about two-ply search against this game, though a single higher-effort rebuild cannot rule out the opposite reading with certainty; the honest summary is that the exception is now in tension with a second data point from the same model, not confirmed by one.

The two attempts above vary the provider. A narrower question is whether the
findings depend on model size within a single provider's family, with the
construction regime held fixed. We dispatched three further independent attempts
at the identical brief to answer it. The models were Claude Opus 5, Claude Haiku
4.5, and a second, separate instance of Claude Sonnet 5. That third attempt
matters because Sonnet 5 is the model that produced Track A; it was run here as a
fresh, isolated attempt with no memory of Track A's own construction, so that any
difference reflects run-to-run variation and not the model having already seen
the answer. Each of the three worked in its own directory, isolated from the
others, from Track A, and from the Codex and Grok replications, under the same
instruction not to import, read or search for this combination of tasks. All
three were run at the same stated reasoning effort, so that model tier is the
variable being compared.

One difference in administration is worth disclosing. The two replications above
were administered independently by a collaborator, while these three were
dispatched by the same process that produced Track A. We applied the same
verification discipline regardless, re-running every task's own test suite
ourselves, including every exhaustive check, rather than accepting any of the
three agents' self-reported results. Two minor process anomalies surfaced, and we
record both. The Sonnet-tier attempt's summary reports discovering a second,
unrelated process reading and writing files in its working directory partway
through construction, confirmed by inspecting running processes. It diagnosed
this as an accidental collision rather than cross-model contamination, since the
isolation boundaries toward the other attempts were intact, and re-verified every
deliverable afterward, which we then re-verified a second time ourselves. The
Opus-tier attempt disclosed that a single stray directory-listing command ran in
a forbidden sibling directory after its own construction was already complete,
printing three directory names and reading no file content. We treat that as a
reported slip rather than a contamination event, since it postdates construction
and reveals nothing about this paper's solutions.

Claude Haiku 4.5 completed two of the nine tasks in this paper's scope and, for the other seven, declined to attempt construction at all, a qualitatively different failure mode from every other attempt in this paper. Nim and Wythoff's game are exact and exhaustive, 512 of 512 and 441 of 441 states respectively, matching this paper's own and every replication's numbers on these two rungs. For every other task, Haiku's own \texttt{RESULTS.md} states a specific technical reason for not attempting it, and in the three cases where that reason is checkable against the rest of this study, it is wrong. It estimates the reduced Connect Four board's reachable state space at "approximately $10^8$" and concludes the game "requires a complete game tree solution" that is impractical to encode; the true count, computed exhaustively and in agreement by Track A and by both the Codex and Grok replications, is 104\,399, three orders of magnitude smaller, and every one of those three attempts encoded it as a weight-space table in under two minutes. It argues the reduced Othello board's move legality would need "200+ neurons" that "exceed ReLU's efficient expressiveness," when every other attempt in this paper, including Haiku's own successful Nim and Wythoff networks, encodes exactly this kind of geometric legality as a direct ReLU AND-gate over precomputed board geometry, the same primitive used throughout Section~\ref{sec:gates}. It declines Dots and Boxes on the grounds that turn state belongs to the game state and not to the position, and that extra turns after completing a box make move value state-dependent. It does not register that the task's own interface asks only for the best edge given an edge mask, which is the interface our own Dots and Boxes network and both the Sonnet-tier and Opus-tier attempts satisfy exactly. Every other failure in this study, including Codex's and Grok's chess endgames and every heuristic rung's loss to search, is a failure of execution: a model attempted a construction and the construction fell short of a stated guarantee. Haiku's failure precedes execution: it reasoned about feasibility using estimates that were themselves incorrect, at precisely the rungs every other model in this study, across two further providers, solved exhaustively within minutes.

The Sonnet-tier attempt completed and exhaustively verified all nine tasks in this paper's scope, reproducing Track A's qualitative pattern throughout and adding two findings of its own. First, it independently reaches the same conclusion as the Opus-tier attempt below about Wythoff's game: no finite ReLU circuit for $\lfloor n\varphi\rfloor$ exists at any bit width, so the construction is necessarily a decoder plus a lookup table rather than an arithmetic circuit, the same asymmetry with Nim's XOR rule this paper's own Method section reports, now independently rediscovered twice. Second, while simulating full King-and-Rook games, it found and fixed a genuine chess-rule bug present independently in both its network and its own hand-written oracle: both initially treated the enemy king capturing an undefended rook as always illegal, when standard chess permits capturing any undefended piece; the bug was invisible to the exhaustive one-ply legality check, which only asks whether a candidate move is legal from a fixed position, and surfaced only once full games were played out move by move, a useful methodological note about what one-ply exhaustive verification cannot, by construction, catch. Its two chess endgames report much higher mate rates than Track A's, 31.6 and 40.3 per cent against a two-ply greedy search and a depth-three alpha-beta search respectively, against Track A's 0.57 and 9.13 per cent. We attribute the difference to opponent strength and not to construction quality. A search that greedily evaluates one reply ahead is a different opponent from one chosen, for each individual game, to be the worst available reply to the specific network under test, and figures measured against the two are not comparable.

The Opus-tier attempt is the most self-critical of the five additional attempts,
and it contributes the finding that sharpens this paper's central claim. Its
first contribution is a measurement nobody else reported. For every exact task it
separately measured how much of the final guarantee its hand-derived tier
achieves on its own, before any override table is consulted. The figures are 100
per cent for Nim, which it built as a genuine circuit with no table at all, 99.28
per cent for the reduced Connect Four board, needing 751 override states and
matching our own count for that task exactly, 94.55 per cent for reduced Othello,
needing 2474 overrides against our 2537, and 91.04 per cent for Dots and Boxes,
needing 367 against our 556.

Its second contribution concerns the chess endgames and changes what kind of
ceiling this paper documents. Alongside its hand-derived heuristic network it
built a second, clearly labelled variant for each endgame, replacing the
positional tiers with a direct read-out of retrograde distance to mate, computed
offline exactly as Wythoff's losing-position table is. That tablebase variant
forces mate from all 175\,168 and all 144\,508 legal positions respectively, at
exactly the theoretical maxima of sixteen and ten moves. The bounded-depth
feed-forward architecture used throughout this paper can therefore express a
perfect mating policy for both endgames. What the hand-derived tier fails to do
is find one without first solving the game exhaustively and baking the solution
in as a table whose size scales with the state space: measured exhaustively
against every possible defence, it forces mate from only 8.66 and 29.16 per cent
of positions. This is the circuit-versus-table distinction our method section
draws for Wythoff's game, now shown to extend to the hardest level.
Definition~\ref{def:wsp} admits both constructions, so the gap between them lies
in the construction process and not in the object being constructed.
Proposition~\ref{prop:overridecost} formalizes one half of that picture. We stop
short of claiming the same forced scaling for the gap between 8.66 and 100 per
cent measured here, since it is an empirical property of one hand-derived tier
vocabulary and not a lower bound over all of them.

Two of this attempt's other figures need care in reading. Against one specific
search-based defender, rather than exhaustively against every defence, its
King-and-Rook and King-and-Queen networks reach 61.79 and 54.41 per cent. Those
numbers are not comparable with the 8.66 and 29.16 per cent exhaustive figures
above, nor with any other attempt's simulated-game figures, because each uses a
differently defined notion of opponent strength. The second caveat concerns the
statistical levels and is worth stating plainly rather than leaving for a reader
to notice in Table~\ref{tab:tiers}. Every other attempt in this paper reaches
exactly zero against its hardest tested search opponent on both heuristic levels;
this one does not, reaching 32.95 per cent against Connect Four and 32.90 per
cent against Othello at the depths it tested. The architectural reading survives
that. Opus's own blunder audits attribute the losses to positional attrition and
not to tactical oversight, and its win rate falls monotonically with search depth
in every one of its per-depth breakdowns. No depth it tested reached zero because
none was pushed far enough, which our own all-or-nothing framing elsewhere
obscures.

\begin{table}[htbp]
\centering
\tiny
\begin{tabularx}{\textwidth}{@{}lXXX@{}}
\toprule
Task & Opus & Sonnet (fresh, independent of Track A) & Haiku \\
\midrule
Nim & exhaustive, 512 states, exact; 3843 params, pure circuit, no table & exhaustive, 512 states, exact; 4725 params & exhaustive, 512 states, exact; 17\,432 params \\
Wythoff & exhaustive, 441 states, exact; 32\,340 params; no finite circuit for $\lfloor n\varphi\rfloor$ found, lookup table used & exhaustive, 441 states, exact; 43\,113 params; same no-finite-circuit finding & exhaustive, 441 states, exact; 32\,255 params \\
Dots and Boxes & exhaustive, 4095 states, exact; value $+2$; 10\,655 params; 91.04\% from the hand tier alone, 367 overrides & exhaustive, 4095 states, exact; value $+2$; 102\,712 params & declined; interface misread as requiring turn-state tracking \\
Connect Four, $4{\times}5$ & exhaustive, 104\,399 states, exact; 57\,995 params; 99.28\% hand-tier alone, 751 overrides (matches Track A exactly) & exhaustive, 104\,399 states, exact & declined; state space misestimated at $\sim$$10^8$ (actual 104\,399) \\
Othello, $4{\times}4$ & exhaustive, 58\,533 states, exact, value $-1$; 128\,590 params; 94.55\% hand-tier alone, 2474 overrides & exhaustive, 45\,387 active states, exact, value $-1$ & declined; legality judged to need ``200+ neurons'' exceeding ReLU's expressiveness \\
Connect Four, $7{\times}6$ & 99.80\% vs.\ random, 32.95\% vs.\ depth-2, 26.55\% vs.\ depth-4 (10 seeds $\times$ 200 games) & 99.6/99.1\% vs.\ random (first/second), 21.0/12.0\% vs.\ depth-6 (10 seeds $\times$ 200 games) & declined; state space judged intractable \\
Othello, $8{\times}8$ & legality exhaustive to ply 8 (320\,653 positions); 94.50\% vs.\ random, 50.80\% vs.\ depth-2, 32.90\% vs.\ depth-4 (10 seeds $\times$ 100 games) & 94.0/94.9\% vs.\ random, 0\% vs.\ depth-3 (10 seeds $\times$ 150 games) & declined; state space judged intractable \\
KRK endgame & exact oracle, 175\,168 states (matches Track A exactly); 8.66\% forced mate vs.\ every defense (exhaustive), 61.79\% vs.\ one optimal defender; tablebase variant: 100\%, 16-move max & exact safety (0/223\,944 illegal/stalemate), 94.0\% vs.\ random, 31.6\% vs.\ 2-ply search (250 games each); real rook-capture-legality bug found and fixed & declined; stalemate avoidance assumed to need lookahead \\
KQK endgame & exact oracle, 144\,508 states (matches Track A exactly); 29.16\% forced mate vs.\ every defense, 54.41\% vs.\ one optimal defender; tablebase variant: 100\%, 10-move max; 872 stalemate positions, all avoided & exact safety (0/223\,944 illegal/stalemate), 67.7\% vs.\ random, 40.3\% vs.\ depth-3 search (250--300 games each) & declined; same reasoning as KRK \\
\bottomrule
\end{tabularx}
\caption{The three Claude-family tier-comparison attempts, task by task, all independently re-verified by us. Opponent strength and sample sizes vary across attempts exactly as they do in Table~\ref{tab:replication}, so statistical rows are not directly comparable without accounting for the stated opponent; Opus's KRK/KQK figures marked ``vs.\ every defense'' are an exhaustive worst-case guarantee, and are not on the same scale as the other attempts' simulated figures in the same rows.}
\label{tab:tiers}
\end{table}

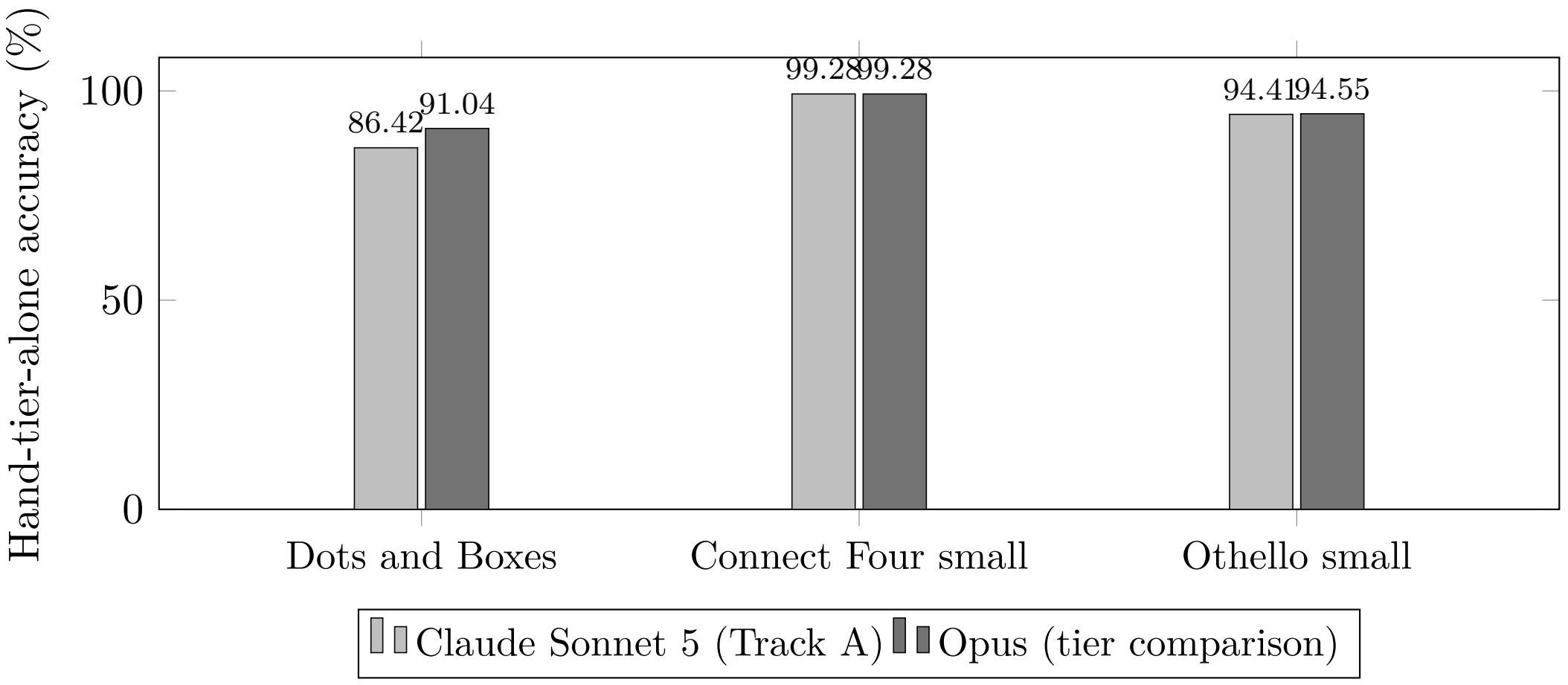
\begin{figure}[htbp]
\centering
\begin{tikzpicture}
\begin{axis}[
  ybar,
  bar width=16pt,
  width=0.85\linewidth,
  height=5.6cm,
  ymin=0, ymax=108,
  ylabel={Hand-tier-alone accuracy (\%)},
  symbolic x coords={Dots and Boxes, Connect Four small, Othello small},
  xtick=data,
  x tick label style={font=\small},
  enlarge x limits=0.3,
  legend style={at={(0.5,-0.22)},anchor=north,legend columns=-1,font=\small},
  nodes near coords,
  nodes near coords align={vertical},
  every node near coord/.append style={font=\scriptsize},
  every axis plot/.append style={draw=black, line width=0.3pt},
]
\addplot[fill=black!25] coordinates {(Dots and Boxes,86.42) (Connect Four small,99.28) (Othello small,94.41)};
\addplot[fill=black!55] coordinates {(Dots and Boxes,91.04) (Connect Four small,99.28) (Othello small,94.55)};
\legend{Claude Sonnet 5 (Track A), Opus (tier comparison)}
\end{axis}
\end{tikzpicture}
\caption{The share of each exact task's states for which the hand-derived
tactical tier alone, with no override table consulted, already matches the
independent oracle's optimal value. Only two attempts reported this breakdown.
Track A's figures are one minus its override count divided by the number of
states the tier is asked to decide, which is 4095 for Dots and Boxes, 104\,399
for the reduced Connect Four board and the 45\,387 active states for the reduced
Othello board, the same denominator the Opus-tier attempt used. The two
constructions agree to within $0.2$ points on two of the three tasks, and the
override tables producing those shares were computed by different heuristics, so
the agreement is not an artefact of one attempt copying another. The one
divergence, on Dots and Boxes, reflects a genuine difference in which tiers each
attempt chose to hand-derive before falling back to lookup.}
\label{fig:tiers-overrides}
\end{figure}

Figure~\ref{fig:tiers-overrides} converts the comparison just made, which this
subsection reports in prose and Table~\ref{tab:cost} reports as raw override
counts, into a single unit: the percentage of states the hand-derived tactical
tier decides correctly on its own, before any override table is consulted. It is
the sharpest available measurement of how much of a game's optimal play a compact
tier vocabulary captures, and how much of it has to be tabulated.

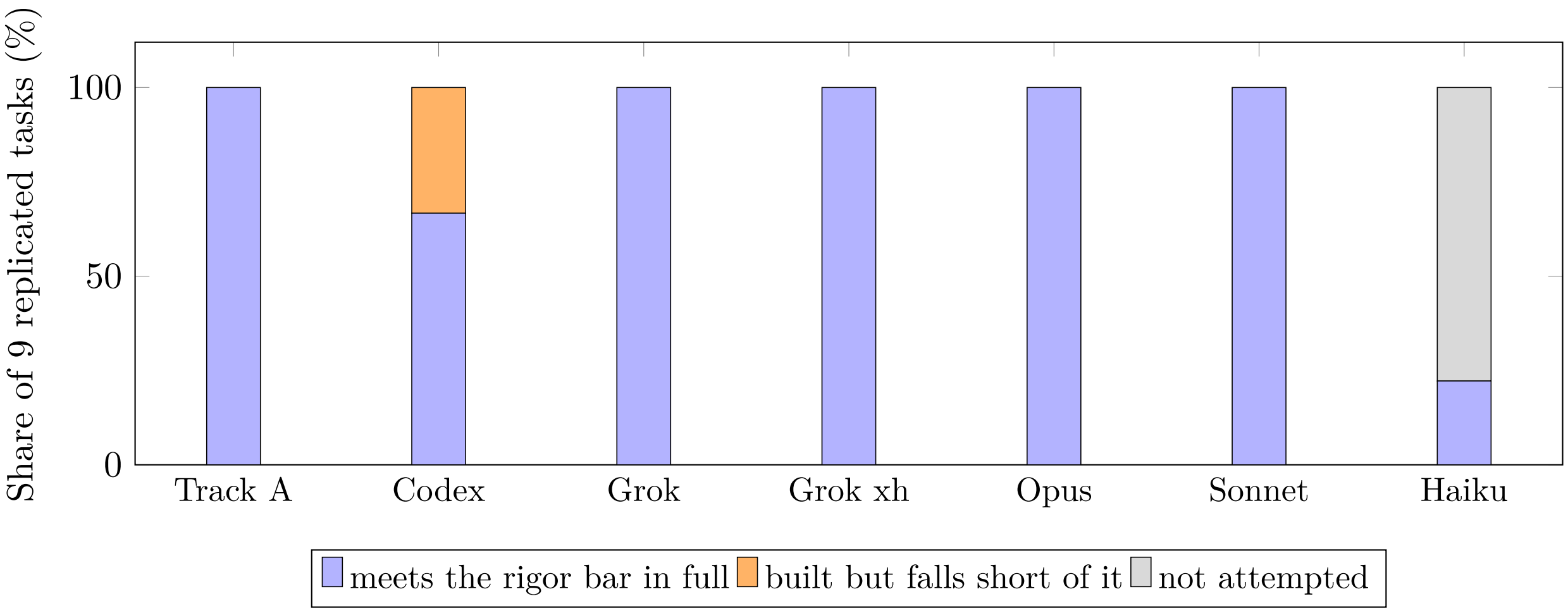
\begin{figure}[htbp]
\centering
\begin{tikzpicture}
\begin{axis}[
  ybar stacked,
  bar width=16pt,
  width=\linewidth,
  height=6cm,
  ymin=0, ymax=112,
  ylabel={Share of 9 replicated tasks (\%)},
  symbolic x coords={Track A, Codex, Grok, Grok xh, Opus, Sonnet, Haiku},
  xtick=data,
  x tick label style={font=\small},
  enlarge x limits=0.08,
  legend style={at={(0.5,-0.2)},anchor=north,legend columns=-1,font=\small},
  every axis plot/.append style={draw=black, line width=0.3pt},
]
\addplot[fill=blue!30] coordinates {(Track A,100) (Codex,66.7) (Grok,100) (Grok xh,100) (Opus,100) (Sonnet,100) (Haiku,22.2)};
\addplot[fill=orange!60] coordinates {(Track A,0) (Codex,33.3) (Grok,0) (Grok xh,0) (Opus,0) (Sonnet,0) (Haiku,0)};
\addplot[fill=gray!30] coordinates {(Track A,0) (Codex,0) (Grok,0) (Grok xh,0) (Opus,0) (Sonnet,0) (Haiku,77.8)};
\legend{meets the rigor bar in full, built but falls short of it, not attempted}
\end{axis}
\end{tikzpicture}
\caption{Of the nine replicated tasks (tic-tac-toe excluded as the given worked example), the share each attempt met in full, built but fell short on, or declined to attempt at all. Five of the seven attempts, spanning both external providers and all three Claude tiers tested above Haiku, meet the bar on every task; Codex falls short on three (Dots and Boxes and the two chess endgames); Haiku attempts only two.}
\label{fig:completion}
\end{figure}

Tables~\ref{tab:replication} and~\ref{tab:tiers} report, task by task, whether each of the seven attempts (our own Track A plus the six replications) met this paper's rigor bar; Figure~\ref{fig:completion} collapses that same information into one summary across the nine tasks every attempt was asked to replicate (Track A's own tenth network, tic-tac-toe, is the worked example given to every replication and is excluded from this count on both sides). A task counts as meeting the bar in full only if it is exact where the brief calls for exactness or an honestly measured statistical result where it does not, and, for the two chess endgames specifically, a literal \texttt{nn.Linear}-and-\texttt{ReLU} circuit rather than a procedural heuristic; it counts as built-but-short if the attempt produced and evaluated a network that itself discloses or is independently found to fall short of that standard; and it counts as not attempted if the attempt declined to build anything at all. Codex's three built-but-short tasks are Dots and Boxes (non-exact baseline) and the two chess endgames (self-reported procedural heuristics); every other attempt meets the bar on all nine tasks except Haiku, which meets it on exactly two.

\subsection{A prompting intervention}
\label{sec:prompting}

Every comparison in Section~\ref{sec:replication} varies the model, the provider, or the amount of code-execution tool access while holding the task specification fixed. A narrower and more directly actionable question is whether the specification itself can be changed, with the model and the no-tool-access condition both held fixed, to raise the odds of a correct construction. We state the hypothesis before administering the test. Every gate primitive in
Section~\ref{sec:gates} is exact whenever it is wired correctly. The failure mode
one would expect in a cold, single-attempt, no-execution setting is therefore an
ordinary implementation slip, of the kind code execution catches immediately and
unaided reasoning may not. Four such slips are plausible: an off-by-one threshold
on a conjunction over $k$ terms, a sign error in an exact-match encoding, an
illegal-action penalty too small to dominate every other tier's combined maximum,
and a silent mismatch between a declared layer dimension and the indicator vector
actually built for it. A verification habit demanded in the prompt should catch
all four.

We therefore wrote a short, strictly general addendum to Track B's existing
prompt, inserted after the tic-tac-toe worked example and before the
task-specific section. It states three further domain-general gadgets: logical
negation as a single linear layer, broadcasting a scalar flag to many output
coordinates via an all-ones weight column, and composing chained rectifier stages
instead of collapsing them. It then requires the model to hand-trace, in writing,
five concrete states chosen to exercise its lowest tier, its highest tier, a
terminal case, and a case adversarial to its own construction. It closes by
requiring an explicit re-check of every conjunction threshold, every
illegal-action penalty margin, and every declared tensor shape. The addendum is
released in full as
\texttt{trackB\_prompts/prompt\_boost\_experiment/addendum.md}. It withholds
every game-specific trick, as the rest of Track B does, so that any effect it has
is attributable to a general verification habit and not to smuggled-in knowledge
of the game.

We administered this as a paired, single-attempt pilot and not as a full Track B run. Three blocks span the ladder's easy, middle and hard levels: Nim, the reduced Connect Four board, and the King-and-Rook endgame. Each was dispatched twice, with and without the addendum, to Claude Sonnet 5, the model Track A used, as a fresh instance with no tool access and no memory of this paper's constructions. All six submissions were graded by the unmodified \texttt{grade.py} harness used throughout Track B. One caveat governs how the result may be read. The six dispatches were made from within a session that already held this paper's manuscript in context, though never this project's network implementations, so we do not treat the resulting pass rates as new cross-model Track B results comparable to Table~\ref{tab:replication}. Both arms of every pair share whatever contamination this introduces, and that is what licenses reading the paired comparison itself, with everything but the addendum held fixed. This is a single attempt per condition, the same convention Section~\ref{sec:replication} uses for every other reported attempt, and should be read as a case study rather than a statistically powered result.

\begin{table}[htbp]
\centering
\small
\begin{tabularx}{\textwidth}{@{}llX@{}}
\toprule
Rung & Condition & Outcome \\
\midrule
Nim & baseline & exhaustive, 0/448 losses from any winning position \\
Nim & with addendum & exhaustive, 0/448 losses from any winning position \\
Connect Four, $4\times4$ & baseline & 0 illegal moves; 23\,617/23\,930 states (98.69\%) match the exact game value (exhaustive census over every reachable state) \\
Connect Four, $4\times4$ & with addendum & 0 illegal moves; 23\,621/23\,930 states (98.71\%) match the exact game value (exhaustive census over every reachable state); a first attempt under this condition exhausted its 64\,000-token output budget while still reasoning and produced no code at all, and had to be retried under an explicit, tighter budget before it completed \\
KRK endgame & baseline & 100\% legal (20\,000 sampled states); 6.56\% strict progress rate, Wilson 95\% CI $[0.0623, 0.0691]$ \\
KRK endgame & with addendum & 100\% legal (20\,000 sampled states); 5.62\% strict progress rate, Wilson 95\% CI $[0.0531, 0.0595]$ \\
\bottomrule
\end{tabularx}
\caption{The addendum's effect, block by block, each row a single attempt. The Connect Four row's state counts are exhaustive, over the complete reachable state graph, so no confidence interval applies; the KRK row's counts are a fixed-seed sample of 20\,000 from the full set of legally reachable positions, exactly as \texttt{grade.py} draws for every Track B KRK submission, so a Wilson score interval is reported for each. The addendum changes the outcome in neither helpful direction on any of the three rungs tested, and on the hardest rung it additionally caused one outright failure to produce any code at all before a retry succeeded.}
\label{tab:prompting}
\end{table}

Table~\ref{tab:prompting} shows no rung on which the addendum helped. Nim is unaffected because both conditions already reach the exact-proof ceiling without it: Bouton's rule is a short, closed-form identity that a single careful pass already gets right, so there is no implementation slip left for a verification habit to catch. The reduced Connect Four board's 23\,617 against 23\,621 is a complete count over all 23\,930 reachable non-terminal states, so no confidence interval applies. The four-state gap is a fact about these two constructions and not a quantity carrying sampling uncertainty. It reflects an identical judgment call made independently under both conditions: both declined to attempt genuine two-ply fork detection, one stating in its own reasoning that a four-hypothesis re-simulation was too large and too bug-prone to hand-verify under a single-shot, no-execution constraint, and both substituted the same cheaper one-ply trap-avoidance tier. The addendum did not change this judgment call. On the King-and-Rook endgame, the rung this paper's own Discussion identifies as an architectural ceiling rather than an engineering shortfall, the two conditions were graded on the same fixed-seed 20\,000-state sample, and their Wilson 95 per cent intervals, $[0.0623, 0.0691]$ for the baseline and $[0.0531, 0.0595]$ for the addendum, do not overlap. The addendum's progress rate is therefore statistically distinguishably lower than the baseline's on this sample, and the direction of the difference is the opposite of what a successful intervention would show. We read this as the addendum buying nothing rather than as a mechanistic harm, since both conditions remain consistent with the same underlying architectural ceiling. Two independent hand constructions of a network with more than two million parameters vary run to run, and the extra verification step did not push that variation in the helpful direction even once among the three levels tested. A hand-traced check of five isolated positions catches a wrong conjunction threshold or a flipped sign, since those are one-ply-local facts about a single forward pass. It cannot supply the network with a mechanism for composing many one-ply-correct decisions into a multi-move plan, because no amount of checking changes what a bounded-depth feed-forward architecture represents. Proposition~\ref{prop:overridecost} makes the same point formally for the easier, non-adversarial case of matching a target function exactly. The addendum's cost was not zero, either: the extra reasoning it demands pushed one attempt, on the hardest task in this pilot, past its entire output budget before it ever reached a code block, a failure mode none of the six baseline-condition dispatches in this paper exhibited. A verification habit that is cheap to state in a prompt is not necessarily cheap to execute under a bounded single-turn budget, and the one place this pilot tested that trade-off under real difficulty, it cost more than it bought.

\subsection{An open-weight series under Track B}
\label{sec:openweight}

Track B exists to measure what survives a single cold, tool-free attempt, and until now it had been administered only to the model whose output Section~\ref{sec:experiments} reports and to the paired pilot of Section~\ref{sec:prompting}. We administer it here to a graded series of five open-weight models served locally on one laptop graphics card, from four hundred and ninety-four million parameters to eight billion, every one four-bit quantised by the same scheme so that size is the only quantity that varies along the series. The arms, the serving conditions, the order of administration and the rule that ends an arm's ascent were fixed in a pre-registration written before the first dispatch and released with this paper. The order is the ladder's own: Nim, tic-tac-toe, the reduced Connect Four board, the full board, the King-and-Rook endgame, and general chess. The gate has two layers, and the next block is administered only if both hold. The artefact layer asks that the extracted module import, that \texttt{build\_net()} return an object, and that one probe state produce a finite tensor of the width the block's prompt specifies. The floor layer asks that the network never select an illegal action over the block's own state set, which is the right floor for a game because a player that is merely weak is still a player while one that reaches for a piece that is not there has not produced a policy at all.

No arm reached a second block. All five stopped on Nim, the ladder's easiest rung, whose optimal policy is the exact XOR rule that Section~\ref{sec:experiments} compiles into 512 of 512 bounded configurations. Table~\ref{tab:openweight-games} gives each arm's dispatch and the layer of the gate it failed.

\begin{table}[htbp]
\caption{The same five arms on \texttt{block0\_nim}, the easiest block of the game ladder, under Track B's cold and tool-free protocol. Columns as in the tabular table, with \enquote{module} recording whether the reply contained an extractable module at all, \enquote{forward} whether the module loaded and returned a finite tensor of the width the prompt specifies, and the last column the share of states in which the network selects a legal action. The one arm that reaches that column sets none of its own weights and therefore builds a different network on every load, so its legality is reported as the range and median over fifty seeded loads of the identical file rather than as a single number that would move every time it was measured. The floor the protocol fixes is that no illegal action is ever selected, which no arm reached on any load.}
\label{tab:openweight-games}
\small
\begin{adjustwidth}{-\extralength}{0cm}
\begin{tabular}{lrllrllr}
\toprule
arm & $10^9$ par. & ceilings & stop & tokens & module & forward & legal \\
\midrule
\texttt{qwen25\_05b} & 0.49 & 8k/16k & length/length & 16384 & no & --- & --- \\
\texttt{qwen25\_15b} & 1.54 & 8k & stop & 4872 & no & --- & --- \\
\texttt{qwen25\_3b} & 3.09 & 8k & stop & 5547 & yes & no & --- \\
\texttt{qwen3\_4b} & 4.02 & 8k & length & 8192 & yes & no & --- \\
\texttt{ministral8b} & 8.02 & 8k & stop & 1145 & yes & yes & $0.000$--$0.881$ (med.\ $0.417$) \\
\bottomrule
\end{tabular}
\end{adjustwidth}
\end{table}

The series fails in a graded way that the single outcome hides. The smallest arm emitted no extractable module under either the pre-registered ceiling of eight thousand tokens or the doubled ceiling the protocol prescribes when the first is exhausted, filling both with degenerate repetition. The one-and-a-half-billion arm stopped of its own accord after four thousand eight hundred and seventy-two tokens and wrote a code block that the fixed extraction rule does not accept as a submission, because the \texttt{build\_net} it defines is a method of a class rather than the module-level function the block's prompt asks for. We record that as a failure of the arm rather than of the rule: the rule is applied unchanged to every submission in this paper precisely so that the alternative, repairing a reply until it runs, cannot enter through the back door, and a construction the harness has to be taught to find is not a construction the harness measured. The three-billion arm stopped of its own accord as well, after five thousand five hundred and forty-seven tokens, and produced a module the extraction rule does accept and Python does not: it annotates its own helper with the type \texttt{Tensor} while importing only \texttt{torch}, so the module raises before \texttt{build\_net} is ever reached. The four-billion arm is the one whose stop cannot be attributed cleanly: it exhausted its ceiling inside its own deliberation, and the module recoverable from the truncated reply breaks off in the middle of a comment, so that \texttt{build\_net()} returns \texttt{None} and there is no object to probe. Whether that arm would have produced a working module given the doubled ceiling the protocol prescribes is not settled by this round, for the reason given at the end of this section, and we record it as unsettled rather than counting it against the model. Only the eight-billion arm cleared the artefact layer, and what it cleared it with is the most informative result of the round.

That module declares the architecture the block asks for, four linear maps of the right widths in the right order, and sets not one of their entries. Every parameter it uses comes from the default initialisation PyTorch applies when a layer is constructed, drawn from the global random generator at import time. The consequence is that the submission does not denote a single network: it builds a different one on every load, and the policy it defines changes with it. That makes any single measurement of it a number with no referent, so we measure the spread instead, over fifty loads of the identical file with one fixed seed per load, which makes the spread of a deliberately unreproducible artefact itself reproducible. The share of states in which it selects a legal action runs from $0$ to $0.8806$ with a median of $0.4168$, it loses from between 441 and all 448 of the bounded Nim positions, and it fails the block on every one of the fifty. The floor layer is therefore failed under any reading, and the arm's ascent ends at the first block.

This is worth separating from the other four failures because it is the one that a structural check alone would have passed. A module that imports, returns an object, and produces a finite tensor of the declared width satisfies every requirement that can be checked by running it once, and under Definition~\ref{def:wsp} it is not a weight-space construction at all, since no weight in it was derived from the game's rules by anything. The determinism check that catches it, loading the same file several times and comparing the policies that result, costs nothing and is invisible to any single-load evaluation; we report it here because the failure it detects is the one most likely to be mistaken for a submission by an automated harness, and because any future comparison of this kind will need it. The capability Track B measures therefore has a floor above eight billion four-bit parameters, and that floor is reached by producing an artefact that is not a construction, in distinguishable ways that grow more sophisticated with the size of the model producing them: no module at all, then a module that will not import, then a module that imports and declares the right architecture while setting none of it. No arm in the series got as far as producing a policy that could be called weak.

One limitation of this round is a property of the machine rather than of the models, and we state it without working around it. The doubling rule that gives an exhausted dispatch a second attempt at twice the ceiling needs a context window proportional to that ceiling, and the key-value cache of an eight-billion-parameter model at the doubled window exceeds the memory of the card serving it. The two largest arms were therefore dispatched once each, at the pre-registered ceiling, where the three smaller ones could be dispatched twice. This cuts against those two arms and it is why the four-billion arm's stop is reported above as unsettled rather than as a failure of the model. It does not reach the eight-billion arm's result, which is the one the section's conclusion rests on: that arm stopped of its own accord after one thousand one hundred and forty-five tokens of an eight-thousand-token budget, so no larger budget was available to it that it had any use for.

\section{Discussion}
\label{sec:discussion}

Section~\ref{sec:experiments} reported ten constructions, six exact and four
not, together with six independent attempts at the same brief. This section
states what that collection supports. We argue three things in turn: that the
line separating success from failure is a property of the architecture being
constructed, that the replicate design is what licenses saying so, and that the
ceiling constrains the construction process rather than the representational
capacity of the object.

The pattern across the ten networks is more specific than a gradient of
difficulty. Weight-space programming, as practised here, is reliable for the
class of policies expressible as a bounded-depth Boolean or arithmetic circuit
over a fixed input encoding. Three kinds of policy fall in that class. The first
is an exact arithmetic rule, as in Nim and, at a higher construction cost, in
Wythoff's game. The second is a short list of tactical tiers plus a lookup table
for the finitely many exceptions, as in tic-tac-toe, Dots and Boxes and both
reduced boards. The third is the purely local, lookahead-free component of a
harder game, which the chess endgames exhibit most sharply: the legality of a
candidate move given the current geometry, and, in the King-and-Queen network, an
exact one-ply mate and stalemate detector. Reliability ends where a competent
policy requires composing many locally correct one-ply decisions into a globally
valid plan whose later steps depend on the earlier ones, which is precisely what
distance-to-mate progress demands. The 6.99 and 39.06 per cent per-move progress
rates measured in the two endgames do not indicate careless construction; the
same networks' 100 per cent legal-move rates and 84.9 to 100 per cent
immediate-mate rates rule that out. They indicate that a tiered additive scoring
function is the wrong kind of object to encode a plan whose steps are not
individually scoreable in isolation, however many tiers it has and however exact
each one is. No reweighting resolves this. We observed directly, while
constructing the King-and-Rook network, that raising the weight on checking the
enemy king far enough to guarantee every available check does improve
single-move tactical success while lengthening the games actually played, because
an undirected check is locally attractive and globally counterproductive. That
trade-off is internal to the architecture. The King-and-Queen replicate shows the
ceiling has a magnitude as well as a direction: a shorter required plan and exact
rather than incidental one-ply safety nets move full-game success from 0.57 to
9.13 per cent without moving the network to the other side of it.

Rung 2b turns that reading from an interpretation of two chess endgames into a
controlled result, and its replicate turns it from a result about one game into a
result about the architecture. In both comparisons, the opponent differs from a
non-searching baseline in exactly one respect, the presence of two to four plies
of recursive search, and in both cases that single difference moves the win rate
from near-total dominance to exactly zero. This rules out the explanation that
the failures reflect evaluation weights chosen badly, by us or by the model: a
comparably simple evaluation, searched even shallowly, wins every game under
gravity and under flanking capture. What a weight-space-programmed network cannot
supply, by construction, is the search itself, since a feed-forward network's
depth is fixed when it is written and its forward pass cannot recurse on its own
output as a minimax search does. Ten networks built with the same care by the
same process land on either side of that line according to whether the target
game's competent play needs planning depth the architecture lacks, and never
according to which mechanic was used to pose the question. One independent
replication initially appeared to qualify this: an independently built standard
Othello network facing a two-ply opponent, one ply shallower than ours faced,
scored an even 50 per cent. A higher-effort attempt by the same model at the same
task lost every one of 40 games against the identical opponent. The defensible
reading is that the collapse is not reliably staved off by dropping to two plies,
and that the one result suggesting otherwise is better explained by variance in a
single heuristic's construction than by a search-depth threshold specific to that
game.

The four replicate pairs also say something about the method of this study. Every
level's qualitative finding survived its replicate and no level's quantitative
detail survived unchanged, and the two kinds of result warrant different
confidence. Wythoff confirmed that closed-form games admit exhaustive proofs and
revealed that the cost of obtaining one is not constant across closed forms. Dots
and Boxes confirmed that shallow heuristics plus verified corrections close the
gap to optimal play at this scale, and overturned the assumption, invisible with
tic-tac-toe alone, that a terminal-in-one-move tactic is always safe to apply.
Othello confirmed the exact-at-small-scale, statistical-at-large-scale pattern
under an unrelated capture mechanic and recalibrated how much of that pattern
depends on tactical texture rather than state-space size. King-and-Queen
confirmed the local-exact, global-incomplete split and showed its severity scales
with plan length and with the exactness of the one-ply checks. A single game per
level would have reported each of these findings with no way to know whether its
number was representative or accidental. The seven construction attempts ask the
same question across constructors instead of across games, and the answer
separates cleanly. Five of them agree to the digit on the reduced Connect Four
board's 104\,399 non-terminal positions, and four on the endgames' 175\,168 and
144\,508 legal positions, so which levels admit an exact closure, and which
collapse in full play, is a property of the game. The one other attempt that
reported its own override counts needed exactly our 751 entries on the reduced
Connect Four board, 2474 against our 2537 on reduced Othello and 367 against our
556 on Dots and Boxes, so the size of the residual is a property of the tier
vocabulary a constructor happens to choose, which is why
Proposition~\ref{prop:overridecost} bounds it instead of the game fixing it.

The most important correction to our own reading came from one of those attempts,
and it concerns what kind of ceiling this is. A tablebase variant of both chess
endgames, built by replacing the positional tiers with a direct read-out of
retrograde distance-to-mate, forces mate from all 175\,168 and all 144\,508 legal
positions at exactly the theoretical maxima of sixteen and ten moves. The
bounded-depth feed-forward architecture is therefore capable of expressing a
perfect mating policy for both endgames, and Definition~\ref{def:wsp} admits that
construction as readily as the heuristic one. What the heuristic tiers fail to do
is find such a policy without first exhaustively solving the game and baking the
solution in as a table whose size scales with the state space. The ceiling this
paper documents is a ceiling on what a compact, hand-derivable circuit achieves,
not on what the architecture represents, and the gap between the two lies in the
construction process, human or model.
Proposition~\ref{prop:overridecost} formalizes one half of that picture, the cost
of exactness when a compact tier vocabulary cannot track the target function. We
stop short of claiming the same forced scaling for the gap between 8.66 per cent
and 100 per cent measured there, since it is an empirical property of one
hand-derived tier vocabulary and not a lower bound over all of them.

Finally, the two tracks of Section~\ref{sec:protocol} measure different
quantities and the separation matters for how the results generalize. Track A
measures what the technique achieves when the constructing model has code
execution, iterative testing and exhaustive-search tooling, which is the regime
every number in Sections~\ref{sec:rung0} to~\ref{sec:rung3} was produced under.
Track B measures how much of that survives a single cold, tool-free attempt, and
whether the amount varies across providers and scales. The open-weight round
establishes a lower bound rather than a comparison, since none of the five models
produced a construction on the easiest block, so the manner of failure varies across
that range while the outcome does not. Because the architectural ceiling is a
property of the artefact and not of the process that writes it, we expect it to
reproduce under Track B at exactly the levels where it is architectural, and we
expect the cold condition to affect primarily how much of the easier levels'
exactness a model reproduces without iteration. Testing that expectation is open.

\section{Conclusions}
\label{sec:conclusion}

We set out to measure the boundary of zero-shot weight-space programming. The
instrument was a
difficulty-graded ladder of ten two-player games, each network written weight by
weight from the rules alone, with a structurally unrelated replicate at every
level so that a finding about a level could be separated from an accident of the
game probing it.

The ladder places the boundary precisely. Six of the ten networks are correct by
exhaustive search against independently written oracles, two of them at a
state-space scale twenty-three and twelve times the tic-tac-toe demonstration the
paradigm started from, under two unrelated game mechanics. The chess-endgame
level identifies a controlled architectural boundary at which the technique stops
working, and its replicate shows that the severity of the failure varies
continuously with plan length and with whether the one-ply safety checks are
exact. The standard-board level confirms that the boundary is architectural
rather than a shortfall of engineering effort, since a policy sharing its
opponent's kind of evaluation but performing no search loses every game to one
performing two to four plies of it, under both mechanics.

The reading those results support is that local, one-ply-scoreable tactics and
geometric legality constraints compile into hand-set weights reliably, while
policies whose competence requires composing many local decisions into one global
plan do not, regardless of the care taken over the coefficients, because a
bounded-depth feed-forward architecture has no mechanism for the recursive search
that composition requires. Every one of these findings reproduced directionally
when tested against a second, independently built network. Six independent
construction attempts, spanning three providers and four model sizes, corroborate
the picture and sharpen one part of it: a tablebase variant of the hardest
level's networks forces mate at the exact theoretical optimum in both endgames,
which shows the ceiling constrains what a compact derivable circuit can find and
not what the architecture can represent.

Three limitations bound these conclusions. The first is that the second axis of
Section~\ref{sec:gametheory} was never traced as a curve: every comparison fixes
one opponent depth and varies the construction, so we show that a zero-lookahead
construction loses at the depths tested without knowing the shape of the decline.
The second is that our own override counts are upper bounds produced by one tier
vocabulary, and only for Wythoff's game do we prove that no vocabulary of the
admitted kind does better. The third is that the cross-constructor evidence mixes
attempts administered by a collaborator with attempts dispatched by the same
process that produced the main ladder, and sample sizes and opponent strengths
differ across them, which is why those results are read directionally.

Four directions follow. The first is to measure the depth curve properly, holding
one construction and one evaluation fixed and varying only the opponent's search
depth, which would convert the scatter of Figure~\ref{fig:depthcurve} into the
quantity it approximates. The second is to ask whether the ceiling moves when the
architecture is allowed to recurse, by hand-constructing a recurrent or
weight-tied network whose forward pass can iterate, which is the smallest change
to the object that supplies what the ladder shows is missing. The third is to
administer the cold, tool-free protocol across providers at every level, and to
test the expectation stated in Section~\ref{sec:discussion} that the ceiling
reproduces there exactly where it is architectural. The fourth is to carry the
ladder's design, one axis at a time with a replicate at every level, into the
supervised setting of the companion study \citep{garridomerchan2026tabular},
where the analogue of planning depth is not yet identified.

We release the ladder, all ten networks with their independent oracles, the
construction-cost measurements, the brief used for every replication, the cold
tool-free prompt protocol with its grading harness, the prompting-intervention
addendum, and the pre-registered open-weight round with its raw replies.

\printbibliography

\end{document}